\documentclass[preprint,5p,12pt,twocolumn]{article}

\usepackage{amssymb}
\usepackage{subfig}
\usepackage{algorithm,algorithmic}
\usepackage{graphicx}
\usepackage{caption}
\usepackage{xspace}
\usepackage{multirow}
\usepackage{rotating}
 \usepackage{url}
\usepackage{gensymb}
\usepackage [ table ]{ xcolor }
\usepackage[top=2cm, bottom=2cm, left=1.5cm, right=1.5cm]{geometry}
\usepackage{fancyhdr}
\usepackage{color}
\usepackage{hyperref}

\usepackage[affil-it]{authblk}

\begin{document}

\pagestyle{fancyplain}

\lhead{\color{red}\textbf{This manuscript is published in Neurocomputing. Please cite it as:}\\
\footnotesize \color{blue}\textit{S.R. Kheradpisheh, M. Ganjtabesh, and T. Masquelier. "Bio-inspired unsupervised learning of visual features leads to robust invariant object recognition." Neurocomputing 205 (2016): 382-392.} \url{http://dx.doi.org/10.1016/j.neucom.2016.04.029}
}




\title{Bio-inspired Unsupervised Learning of Visual Features Leads to Robust Invariant Object Recognition}


\author{Saeed Reza Kheradpisheh$ ^{1,5} $ }
\author{Mohammad Ganjtabesh$ ^{1,} $\thanks{Corresponding author.\\ Email addresses:\\ kheradpisheh@ut.ac.ir (SRK),\\ mgtabesh@ut.ac.ir (MG),\\ timothee.masquelier@alum.mit.edu (TM).}}
\author{Timoth\'ee Masquelier$ ^{2,3,4,5} $}
\affil{\footnotesize $ ^{1} $ School of Mathematics, Statistics, and Computer Science, University of Tehran, Tehran, Iran\\ $ ^{2} $ INSERM, U968, Paris, F-75012, France\\ $ ^{3} $ Sorbonne Universit\'es, UPMC Univ Paris 06, UMR-S 968, Institut de la Vision, Paris, F-75012, France \\  $ ^{4} $ CNRS, UMR-7210, Paris, F-75012, France\\ $ ^{5} $ CERCO UMR 5549, CNRS – Universit\'e de Toulouse, F-31300, France}
\date{}
\maketitle
\begin{abstract}
Retinal image of surrounding objects varies tremendously due to the changes in position, size, pose, illumination condition, background context, occlusion, noise, and nonrigid deformations. But despite these huge variations, our visual system is able to invariantly recognize any object in just a fraction of a second. To date, various computational models have been proposed to mimic the hierarchical processing of the ventral visual pathway, with limited success. Here, we show that the association of both biologically inspired network architecture  and learning rule significantly improves the models' performance when facing challenging invariant object recognition problems. Our model is an asynchronous feedforward spiking neural network. When the network is presented with natural images, the neurons in the entry layers detect edges, and the most activated ones fire first, while neurons in higher layers are equipped with spike timing-dependent plasticity. These neurons progressively become selective  to intermediate complexity visual features appropriate for object categorization. The model is evaluated on \emph{3D-Object} and \emph{ETH-80} datasets which are two benchmarks for invariant object recognition, and is shown to outperform state-of-the-art models, including DeepConvNet and HMAX. This demonstrates its ability to accurately recognize different instances of multiple object classes even under various appearance conditions (different views, scales, tilts, and backgrounds). Several statistical analysis techniques are used to show that our model extracts class specific and highly informative features.

\textbf{Keywords:} View-Invariant Object Recognition, Visual Cortex, STDP, Spiking Neurons, Temporal Coding

\end{abstract}






\section{Introduction}
\label{intro}
Humans can effortlessly and rapidly recognize surrounding objects~\cite{thorpe1996speed}, despite the tremendous variations in the projection of each object on the retina~\cite{biederman1987recognition} caused by various transformations such as changes in object position, size, pose, illumination condition and background context~\cite{dicarlo2012does}. This invariant recognition is presumably handled through hierarchical processing in the so-called ventral pathway. Such hierarchical processing starts in V1 layers, which extract simple features such as bars and edges in different orientations~\cite{lennie2005coding}, continues in  intermediate layers such as V2 and V4, which are responsive to more complex features~\cite{nandy2013fine}, and culminates in the inferior temporal cortex (IT), where the neurons are selective to object parts or whole objects~\cite{tanaka1991coding}. By moving from the lower layers to the higher layers, the feature complexity, receptive field size and transformation invariance increase, in such a way that the IT neurons can invariantly represent the objects in a linearly separable manner~\cite{hung2005fast,rust2010selectivity}. 

Another amazing feature of the primates' visual system is its high processing speed. The first wave of image-driven neuronal responses in IT appears around 100 ms after the stimulus onset~\cite{thorpe1996speed,dicarlo2012does}. Recordings from monkey IT cortex have demonstrated that the first spikes (over a short time window of 12.5 ms), about 100 ms after the image presentation,  carry accurate information about the nature of the visual stimulus~\cite{hung2005fast}. Hence,  ultra-rapid object recognition is presumably performed in a feedforward manner~\cite{dicarlo2012does}. Moreover, although there exist various intra- and inter-area feedback connections in the visual cortex, some neurophysiological~\cite{liu2009timing, freiwald2010functional,dicarlo2012does} and theoretical~\cite{Anselmi2014} studies have also suggested that the feedforward information is usually sufficient for invariant object categorization.  

Appealed by the impressive speed and performance of the primates' visual system, computer vision scientists have long tried to ``copy'' it. So far, it is mostly the architecture of the visual system that has been mimicked. For instance, using hierarchical feedforward networks with restricted receptive fields, like in the brain, has been proven useful~\cite{Fukushima1980,LeCun1998,Serre2007.PAMI,Lee2009,Krizhevsky2012,Le2013}. In comparison, the way that biological visual systems learn the appropriate features has attracted much less attention. All the above-mentioned approaches somehow use non biologically plausible learning rules. Yet the ability of the visual cortex to wire itself,  mostly in an unsupervised manner, is remarkable~\cite{Ghose2004,Kourtzi2006}. 
\headheight=0pt
\lhead{}
\pagestyle{plain}

 Here, we propose that adding bio-inspired learning to bio-inspired architectures could improve the models' behavior. To this end,
we focused on a particular form of synaptic plasticity known as spike timing-dependent plasticity (STDP), which has been observed in the mamalian visual cortex~\cite{Meliza2006,Huang2014}. Briefly, STDP reinforces the connections with afferents that significantly contributed to make a neuron fire, while it depresses the others~\cite{Feldman2012}. A recent psychophysical study provided some indirect evidence for this form of plasticity in the human visual cortex~\cite{McMahon2012}.

In an earlier study~\cite{Masquelier2007}, it is shown that a combination of a temporal coding scheme --  where in the entry layer of a spiking neural network  the most strongly activated neurons fire first -- with STDP leads to a situation where neurons in higher visual areas will gradually become selective to complex visual features in an unsupervised manner. These features are both salient and consistently present in the inputs.  Furthermore, as learning progresses, the neurons' responses rapidly accelerates. These responses can then be fed to a classifier to do a categorization task.

In this study, we show that such an approach strongly outperforms state-of-the-art computer vision algorithms on view-invariant object recognition benchmark tasks including  3D-Object~\cite{4408987,Pepik2014} and ETH-80~\cite{leibe2003analyzing} datasets. These datasets contain natural and unsegmented images, where objects have large variations in scale, viewpoint, and tilt, which makes their recognition hard~\cite{Pinto2008}, and probably out of reach for most of the other bio-inspired models~\cite{pinto2011comparing,ghodrati2014feedforward}.  Yet our algorithm generalizes surprisingly well, even when ``simple classifiers'' are used, because STDP naturally extracts features that are class specific. This point was further confirmed using mutual information~\cite{pohjalainen2013feature} and representational dissimilarity matrix (RDM)~\cite{kriegeskorte2008representational}. Moreover, the distribution of objects in the obtained feature space was analyzed using hierarchical clustering~\cite{murtagh2012algorithms}, and objects of the same category tended to cluster together.

\begin{figure*}[!htb]
\centering
\includegraphics[scale=0.45]{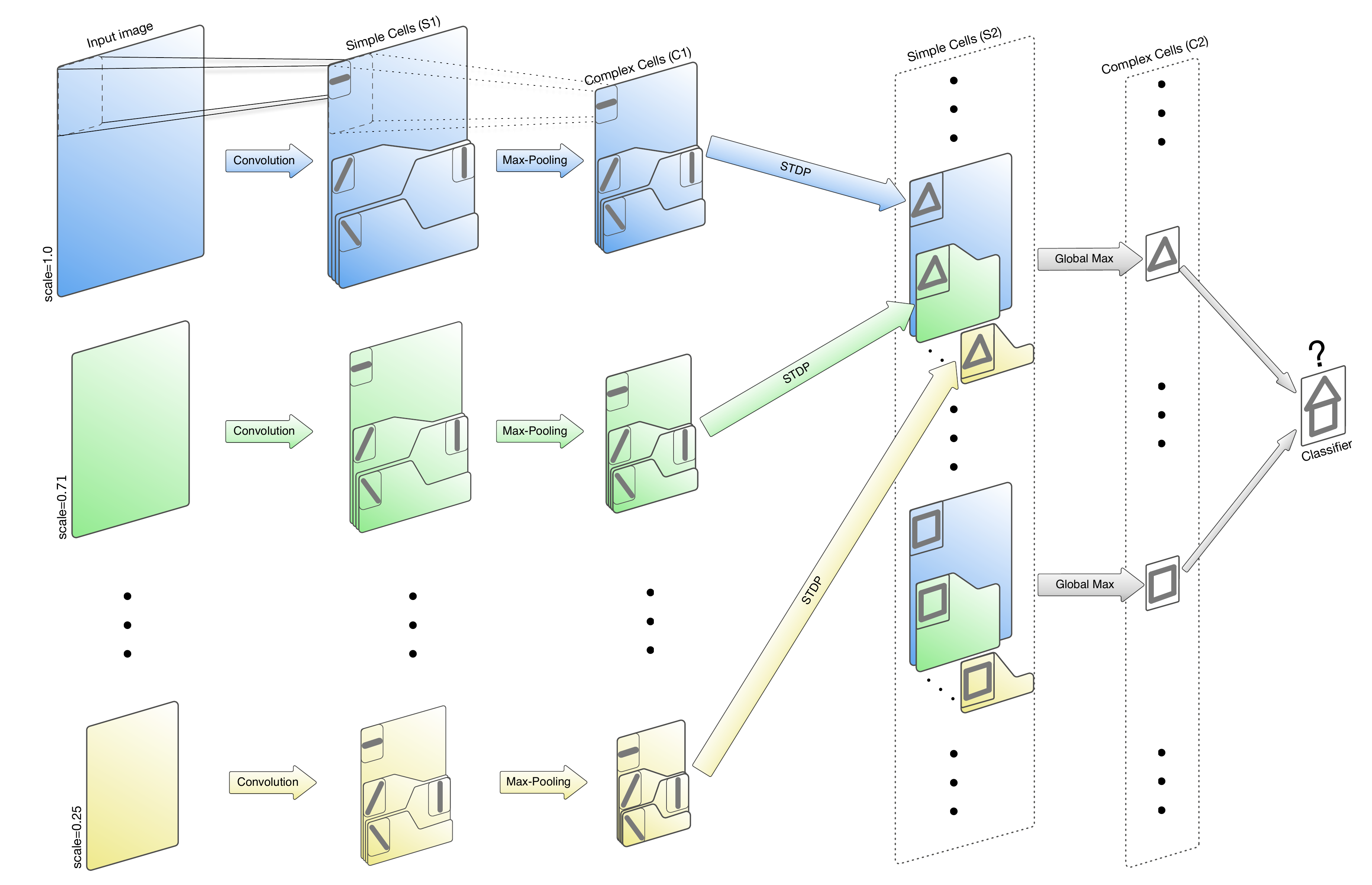}
\caption{ Overview of our 5 layered feedforward spiking neural
network. The network processes the input image in a multi-scale form, each processing scale is shown with a different color.
Cells are organized in retinotopic maps until the $S_2$ layer (included). $S_1$ cells of each processing scale detect edges from the corresponding scaled image. $C_1$ maps sub-sample the corresponding $S_1$
maps by taking the maximum response over a square neighborhood.
$S_2$ cells are selective to intermediate complexity visual
features, defined as a combination of oriented edges of a same scale(here we
symbolically represented a triangle detector and a square detector).
There is one $S_1$--$C_1$--$S_2$ pathway for each processing scale. Then $C_2$ cells take the maximum
response of $S_2$ cells over all positions and scales and are thus
shift and scale invariant. Finally, a classification is done based
on the $C_2$ cells' responses (here we symbolically represented a
house/non-house classifier). $C_1$ to $S_2$ synaptic connections are learned with
STDP, in an unsupervised manner.
} \label{model_figure}
\end{figure*}

\section{Materials and methods}\label{MaterialsAndMethods}

The algorithm we used here is a scaled-up version of the one presented in~\cite{Masquelier2007}. Essentially, many more C2 features and iterations were used. Our code is available upon request. We used a five-layer hierarchical network $S_1$ $\rightarrow$ $C_1$ $\rightarrow$ $S_2$ $\rightarrow$ $C_2$ $\rightarrow$ $classifier$, largely inspired by the HMAX model~\cite{Serre2007.PAMI} (see Fig.~\ref{model_figure}). Specifically, we alternated simple cells that gain selectivity through a sum operation, and complex cells that gain shift and scale invariance through a max operation. However, our network uses spiking neurons and operates in the temporal domain: when presented with an image, the first layer's $S_1$ cells, detect oriented edges and the more strongly a cell is stimulated the earlier it fires. These $S_1$ spikes are then propagated asynchronously through the feedforward network. We only compute the first spike fired by each neuron (if any), which leads to efficient implementations. The justification for this is that later spikes are probably not used in ultra-rapid visual categorization tasks in primates~\cite{Thorpe1989}. We used restricted receptive fields and a weight sharing mechanism (i.e. convolutional network). In our model, images are presented sequentially and the resulting spike waves are propagated through to the $S_2$ layer, where STDP is used to extract diagnostic features. 

More specifically, the first layer's $S_1$ cells detect bars and edges using Gabor filters. Here we used $5\times 5$ convolutional kernels corresponding to Gabor filters with the wavelength of 5 and four different preferred orientations ($\pi/8,\pi/4+\pi/8 , \pi/2+\pi/8, 3\pi/4+\pi/8$).  These filters are applied to five scaled versions of the original image: 100\%, 71\%, 50\%, 30\%, and 25\% (each processing scale declared by a different color in Fig.~\ref{model_figure}). Hence, for each scaled version of the input image we have four $S_1$ maps (one for each orientation), and overall, there are 4$\times$5 $=$ 20 maps of $S_{1}$ cells (see the $S_{1}$ maps of Fig.~\ref{model_figure}). Evidently, the $S_1$ cells of larger scales detect edges with higher spatial frequencies while the smaller scales extract edges with lower spatial frequencies. Indeed, instead of changing the size and spatial frequency of Gabor filters, we are changing the size of input image. This is a way to implement scale invariance at a low computational cost.

Each $S_{1}$ cell emits a spike with a latency that is inversely proportional to the absolute value of the convolution. Thus, the more strongly a cell is stimulated the earlier it fires (intensity-to-latency conversion, as observed experimentally~\cite{Celebrini1993,Albrecht2002,Shriki2012}). To increase the sparsity at a given scale and location (corresponding to one cortical column), only the spike corresponding to the best matching orientation is propagated (i.e. a winner-take-all inhibition is employed). In other word, for each position in the four $S_1$ orientation maps of a given scale, the $S_1$ cell with highest convolution value emits a spike and prevents the other three $S_1$ cells from firing.

For each $S_{1}$ map, there is a corresponding $C_{1}$ map. Each $C_{1}$ cell propagates the first spike emitted by the $S_{1}$ cells in a $7\times7$ square neighborhood of the $S_{1}$ map which corresponds to one specific orientation and one scale (see the $C_{1}$ maps of Fig.~\ref{model_figure}). $C_{1}$ cells thus execute a maximum operation over the $S_{1}$ cells with the same preferred feature across a portion of the visual field, which is a biologically plausible way to gain local shift invariance~\cite{Riesenhuber1999,Rousselet2003}. The overlap between the afferents of two adjacent $C_1$ cells is just one $S_1$ row, hence a subsampling over the $S_1$ maps is done by the $C_1$ layers as well. Therefore, each $C_1$ map has $6\times6=36$ fewer cells than the corresponding $S_1$ map.

$S_{2}$ features correspond to intermediate-complexity visual features which are optimum for  object classification~\cite{Ullman2002}. Each $S_{2}$ feature has a prototype $S_{2}$ cell (specified by a $C_{1}$-$S_{2}$ synaptic weight matrix), which is a weighted combination of bars ($C_{1}$ cells) with different orientations in a $16 \times 16$ square neighborhood. Each prototype $S_{2}$ cell is retinotopically duplicated in the five scale maps (i.e. weight-sharing is used). Within those maps, the $S_{2}$ cells can integrate spikes only from the
four $C_{1}$ maps of their corresponding processing scales. This way, a given $S_{2}$ feature is simultaneously explored in all positions and scales (see $S_{2}$ maps of Fig.~\ref{model_figure} with same feature prototype but in different processing scales specified by different colors). Indeed, duplicated cells in all positions of all scale maps integrate the spike train in parallel and compete with each other. The first duplicate reaching its threshold, if any, is the winner. The winner fires and prevents the other duplicated cells in all other positions and scales from firing through a winner-take-all inhibition mechanism. Then, for each prototype, the winner $S_{2}$ cell triggers the unsupervised STDP rule and its weight matrix is updated. The changes in its weights are applied over all other duplicate cells in different positions and scales (weight sharing mechanism). This allows the system to learn frequent patterns, independently of their position and size in the training images. 

The learning process begins with $S_{2}$ features initialized by random numbers drawn from a normal distribution with mean $0.8$ and STD $0.05$, and the threshold of all $S_{2}$ cells is set to 64 ($= 1/4 \times 16 \times 16$). Through the learning process, a local inhibition between different $S_{2}$ prototype cells is used to prevent the convergence of different $S_{2}$ prototypes to similar features: when a cell fires at a given position and scale, it prevents all the other cells (independently of their preferred prototype) from firing later at the same scale and within a neighborhood around the firing position. Thus, the cell population self-organizes, each cell trying to learn a distinct pattern so as to cover the whole variability of the inputs. Moreover, we applied a k-winner-take-all strategy in $S_{2}$ layer to ensure that at most two cells can fire for each processing scale. This mechanism, only used in the learning phase, helps the cells to learn patterns with different real sizes. Without it, there is a natural bias toward ``small" patterns (i.e., large scales), simply because corresponding maps are larger, and so likeliness of firing with random weights at the beginning of the STDP process is higher.

A simplified version of STDP is used to learn the $C_{1}-S_{2}$ weights as follows:

  $$\left\{ 
  \begin{array}{l l l}
    \Delta w_{ij}=a^{+}.w_{ij}.(1-w_{ij}),&  if & t_{j}-t_{i}\leq 0, \\
    \Delta w_{ij}=a^{-}.w_{ij}.(1-w_{ij}),&  if & t_{j}-t_{i} > 0,
  \end{array} \right.$$  
where $ i $ and $ j $ respectively refer to the index of post- and presynaptic neurons,
$t_i$ and $t_j$ are the corresponding spike times, $\Delta w_{ij}$ is the synaptic weight
modification, and $a^{+}$ and $a^{-}$ are two parameters specifying the
learning rate. Note that the exact time difference between two spikes ($t_{j}-t_{i}$) does not affect the weight change, but only its sign is considered. These simplifications are equivalent to assuming that the intensity-to-latency conversion of $S_{1}$ cells compresses the whole
spike wave in a relatively short time interval (say, $20-30$ ms), so that all presynaptic spikes necessarily fall close to the postsynaptic spike time, and the time lags are negligible. The multiplicative term $w_{ij}.(1-w_{ij})$ ensures the weights remain in the range [0,1] and maintains all synapses in an excitatory mode. The learning phase starts by $a^{+}=2^{-6}$ which is multiplied by 2 after each 400 postsynaptic spikes up to a maximum value of $2^{-2}$. A fixed $a^{+}/a^{-}$ ratio (-4/3) is used. This allows us to speed up the convergence of $S_{2}$ features as the learning progresses. Initiation of the learning phase with high learning rates would lead to erratic results.

For each $S_{2}$ prototype, a $C_{2}$ cell propagates the first spike emitted by the corresponding $S_{2}$ cells over all positions and processing scales, leading to the global shift- and scale-invariant cells (see the $C_{2}$ layer of Fig.~\ref{model_figure}).

\section{Experimental Results}\label{Results}
\subsection{Dataset and Experimental Setup}
\begin{figure*}
\centering
\includegraphics[scale=0.9]{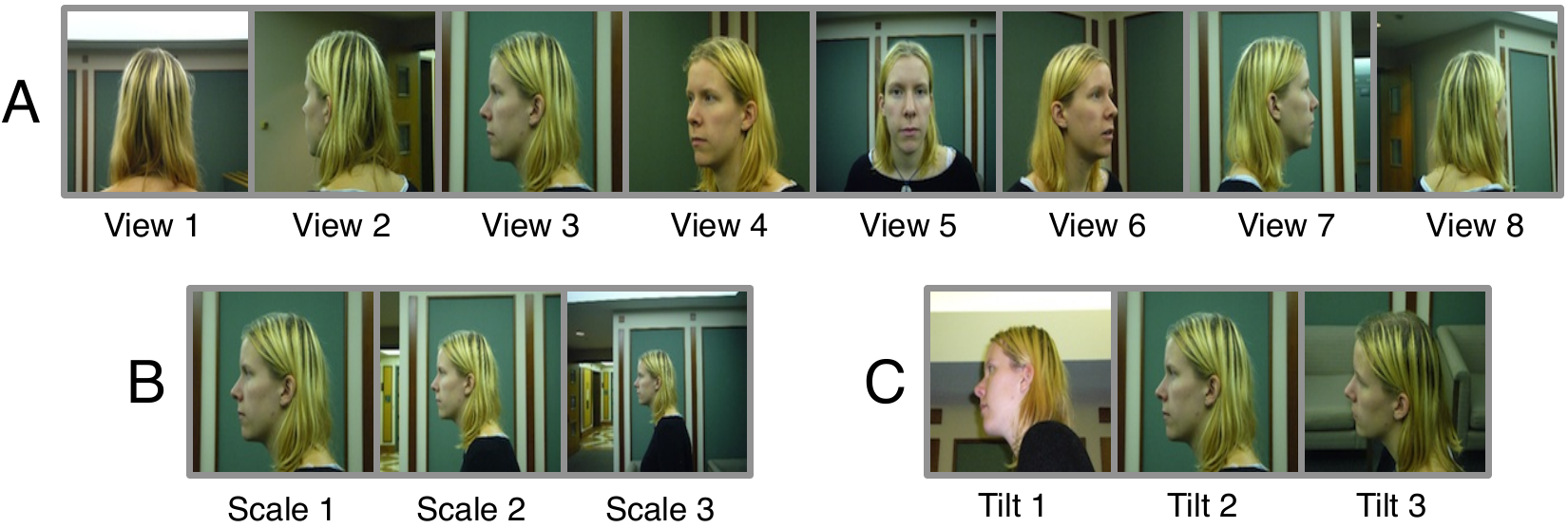}
\caption{Some images of the head class of 3D-Object dataset in different A) views, B) scales, and C) tilts.}
\label{objects}
\end{figure*}

To study the robustness of our model with respect to different transformations such as scale and viewpoint, we evaluated it on the \emph{3D-Object} and \emph{ETH-80} datasets. The 3D-Object is provided by Savarese et al. at CVGLab, Stanford University~\cite{4408987}. This dataset contains 10 different object classes: bicycle, car, cellphone, head, iron, monitor, mouse, shoe, stapler, and toaster. There are about 10 different instances for each object class. The object instances are photographed in about 72 different conditions: eight view angles, three distances (scales), and three different tilts. The images are not segmented and the objects are located in different backgrounds (the background changes even for different conditions of the same object instance). Figure~\ref{objects} presents some examples of objects in this dataset.  

The ETh-80 dataset includes 80 3D objects in eight different object categories including apple, car, toy cow, cup, toy dog, toy horse, pear, and tomato. Each object is photographed in  41 viewpoints with different view angles and different tilts. Figure S1 in Supplementary Information provides some examples of objects in this dataset from different viewpoints.

For both datasets, five instances of each object category are selected for the training set to be used in the learning phase. The remaining instances constitute the testing set which is not seen during the learning phase, but is used afterward to evaluate the recognition performance. This standard cross-validation procedure allows to measure the generalization ability of the model beyond the specific training examples. Note that for 3D-Object dataset, the original size of all images were preserved, while the images of ETH-80 dataset are resized to $300$ pixels in height while preserving the aspect ratio. The images of both datasets were converted to grayscale values.

\begin{table*}
\caption{Performance of our model, HMAX, and DeepConvNet with different number of features.}
\label{table_example1}
\scriptsize
\centering
\begin{tabular}{|c|c|c|c|c|c||c|c|c|c||c|c|}
\hline 
Dataset& &  \multicolumn{4}{c||}{Our model}&\multicolumn{4}{c||}{HMAX}&DeepConvNet \\ 
\hline
\hline
\multirow{2}{*}{3D-Object}&\# Features & 200 & 300 & 400 & 500&1000&3000&9000&12000&4096\\
\cline{2-11}
&Accuracy &  76.1\% & 94.7\% & 96.0\% & 96.0\%&58.2\%&60.1\%&61.9\%&62.4\%&85.8\%\\
\hline \hline
\multirow{2}{*}{ETH-80}&\# Features &  500 & 750 & 1000& 1250&500&1000&2000&5000&4096\\
\cline{2-11}
&Accuracy &  75.3 & 79.3\% & 80.7\% & 81.1\%&66.3\%&68.7\%&68.9\%&69.0\%&79.1\%\\
\hline
\end{tabular}
\end{table*}
As already mentioned, the building process of $S_2$ features is performed in a completely unsupervised manner. Hence, through the execution of the unsupervised STDP-based learning, the training images are randomly fed into the model (without considering their class labels, viewpoints, scales, and tilts). The learning process starts with initial random weights and finishes when 600 spikes have occurred in each $S_2$ map. Then STDP is turned off, and the ability of the obtained features  to invariantly represent different object classes is evaluated. To compute the corresponding $C_2$ feature vector for each input image, the thresholds of $C_2$ neurons are set to infinite, and their final potentials are evaluated, after propagating the whole spike train generated by each image. Each final potential can be seen as the number of early spikes in common between the current input and a stored prototype (this is very similar to the tuning operation of $S$ cells in HMAX). Then, a one-versus-one multiclass linear support vector machine (SVM) classifier is trained based on the $C_2$ features of the training set and it is evaluated on the test set. 

We have compared the performance of our model with the HMAX model~\cite{Serre2007.PAMI} and deep supervised convolutional network (DeepConvNet) by Krizhevsky et. al~\cite{Krizhevsky2012}. Comparison with the HMAX model is particularly instructive, since as explained above, we use very similar architecture, tuning and maximum operations. The main difference is that instead of using an unsupervised learning rule like us, the HMAX model uses random crops from the training images to imprint the $S2$ features (here of equal size). Then a SVM classifier was trained over the HMAX $C2$ features to complete the object recognition process. The employed HMAX model is implemented by Mutch, et al.~\cite{MUTCH10} and the codes are publicly available at http://cbcl.mit.edu/jmutch/cns/index.html.

We also compared our model with DeepConvNet which has been shown to be the best algorithm in various object classification tasks including the ImageNet LSVRC-2010 contest~\cite{Krizhevsky2012}. It is comprised of eight consecutive layers (five convolutional layers followed by three fully connected layers) with about 60 millions parameters, learned with stochastic gradient descent. We have used a pre-trained DeepConvNet model implemented by Jia, et al.~\cite{jia2014caffe}, whose code is also available at http://caffe.berkeleyvision.org. The training was done over the ILSVRC2012 dataset (a subset of ImageNet) with about 1.2 million images in 1000 categories. We fed the training and testing images into DeepConvNet and extracted the feature values from the 7th layer. Again, a SVM is used to do the object recognition based on the extracted features.
\subsection{Performance analysis}
Table~\ref{table_example1} provides the accuracy of our model in category classification independently of view, tilt, and scale, when different number of $S_2$ features are learned by the STDP-based learning algorithm. The  results indicate that the model reaches a high classification performance on 3D-Object dataset with about 300 $C_2$ features only (about 30 features per class). The performance is flattened around 96\% for feature vectors of size greater than 400. Also, for the ETH-80 dataset, the model attains to a reasonable recognition accuracy of about 81\% with only 1250 extracted features.  We have also performed the same experiments on HMAX and DeepConvNet models which their accuracies are also provided in Table~\ref{table_example1}.

Performance of the HMAX model was weak on both datasets, which is not too surprising, several previous studies have shown that the performance of the HMAX model extensively decreases when facing significant object transformations~\cite{Pinto2008,pinto2011comparing}. Given the structural similarities between our model and HMAX, the superiority of our model is presumably related to the unsupervised feature learning. Indeed, most of the randomly extracted S2 patches in HMAX tend to be redundant and irrelevant, as we will see in the next section.  

DeepConvNet reached a mean performance of about 86\% on 3D-Object and about 79\% on ETH-80 dataset.  Thus, our model outperforms DeepConvNet on both datasets, which itself significantly outperforms HMAX. It should be noted that the images of each object in these two datasets are highly varied (e.g., in 3D-Object dataset, there is a 45\degree difference between two successive views of an object) and it has previously been shown that the performance of DeepConvNet drops when facing such transformations~\cite{ghodrati2014feedforward}. Another drawback of DeepConvNet is that, due to the large number of parameters, it needs to be trained over millions of images to avoid overfitting~\cite{cox2014neural} (here we avoided this problem by using a pre-trained version, but doing the training on {about 3500} images we used here would presumably lead to massive overfitting). Conversely,  our model is able to learn objects using much fewer images.


Consequently, the results indicate that our model has a great ability to learn diagnostic features tolerating transformations and deformations of the presented stimulus.

\subsection{Feature Analysis}
\begin{figure*}
\centering
\subfloat[]{\includegraphics[width=2.2in]{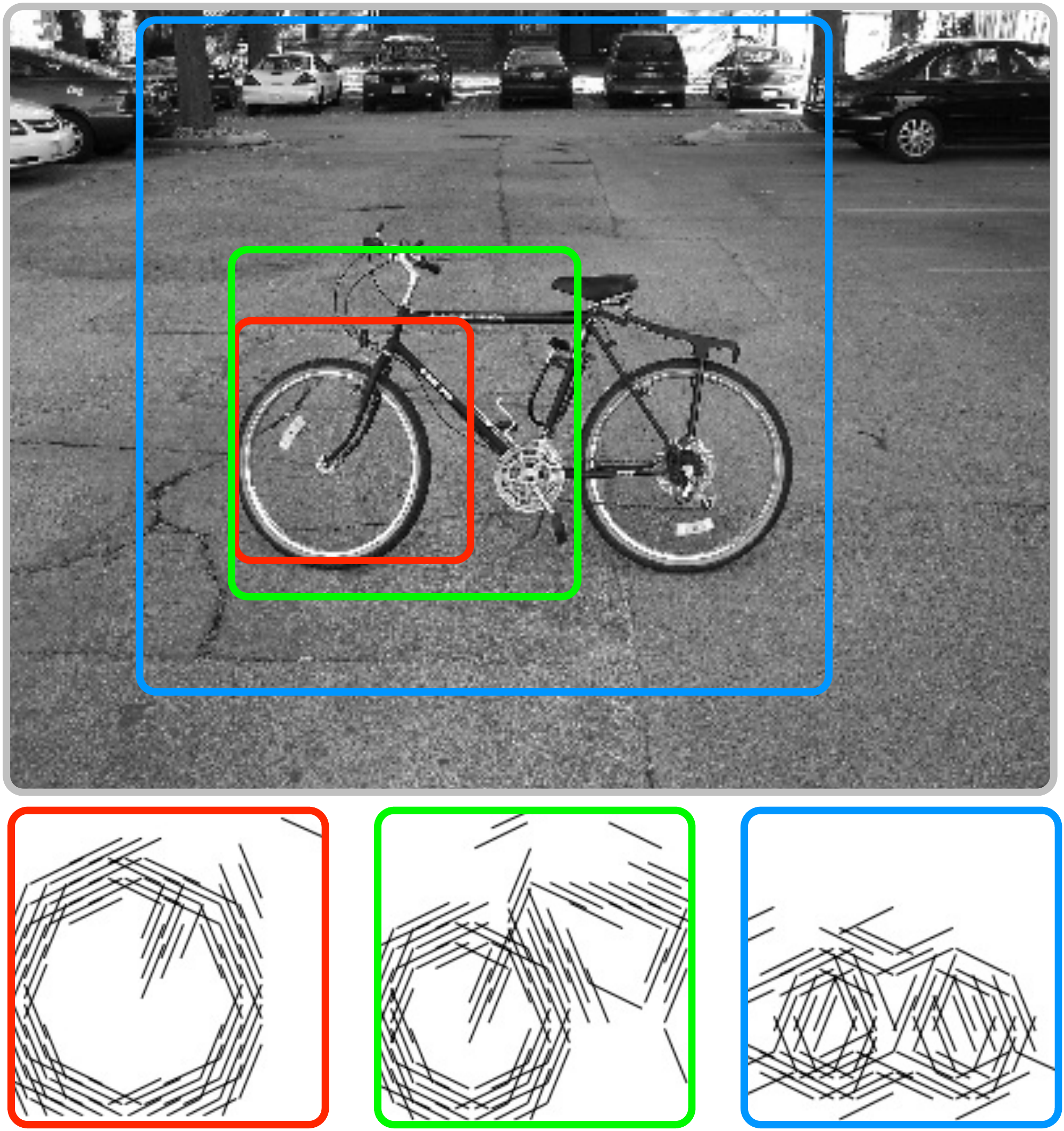}%
\label{fig_first_case}}
\hfil
\subfloat[]{\includegraphics[width=2.2in]{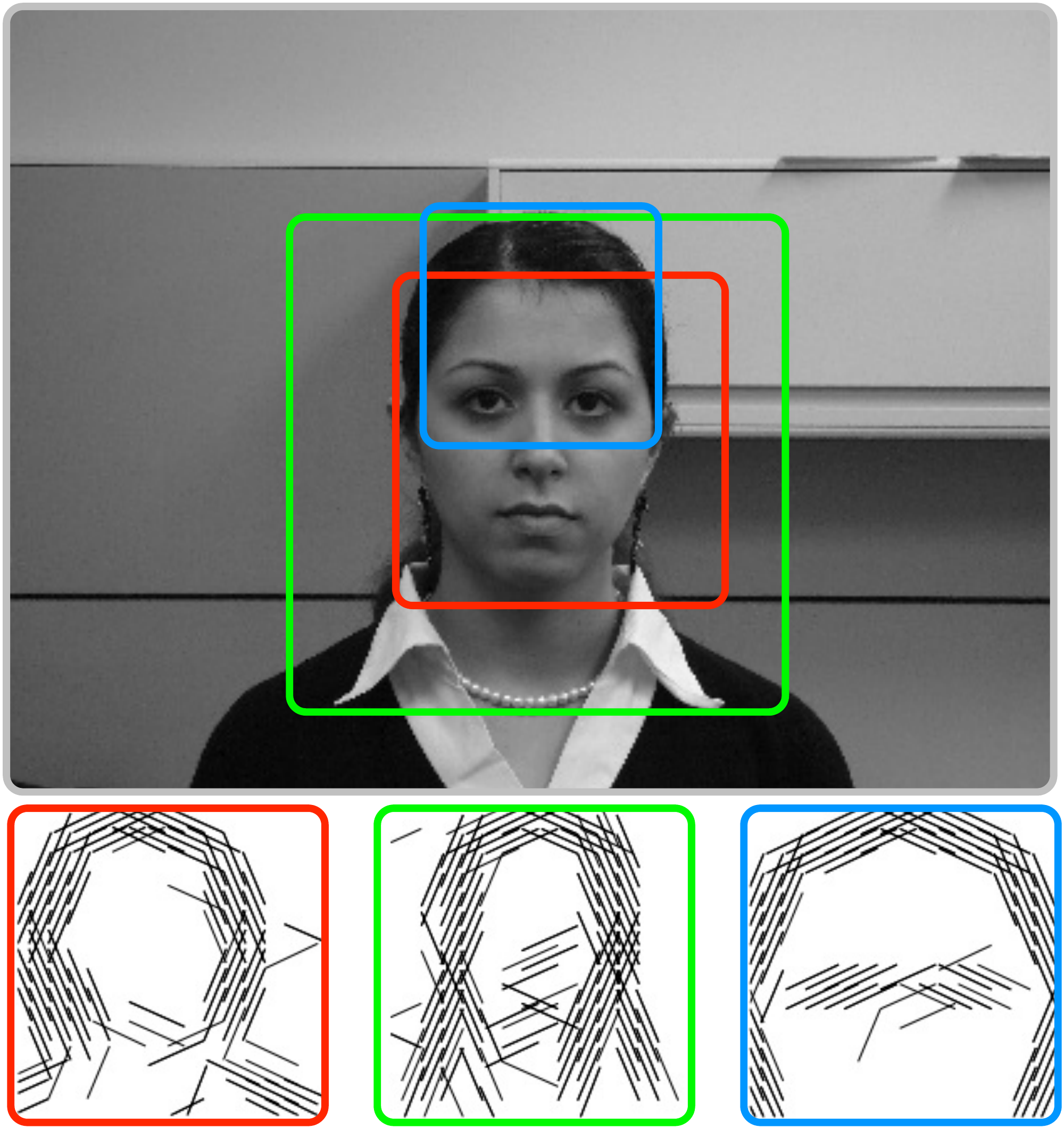}%
\label{fig_second_case}}
\hfil
\subfloat[]{\includegraphics[width=2.2in]{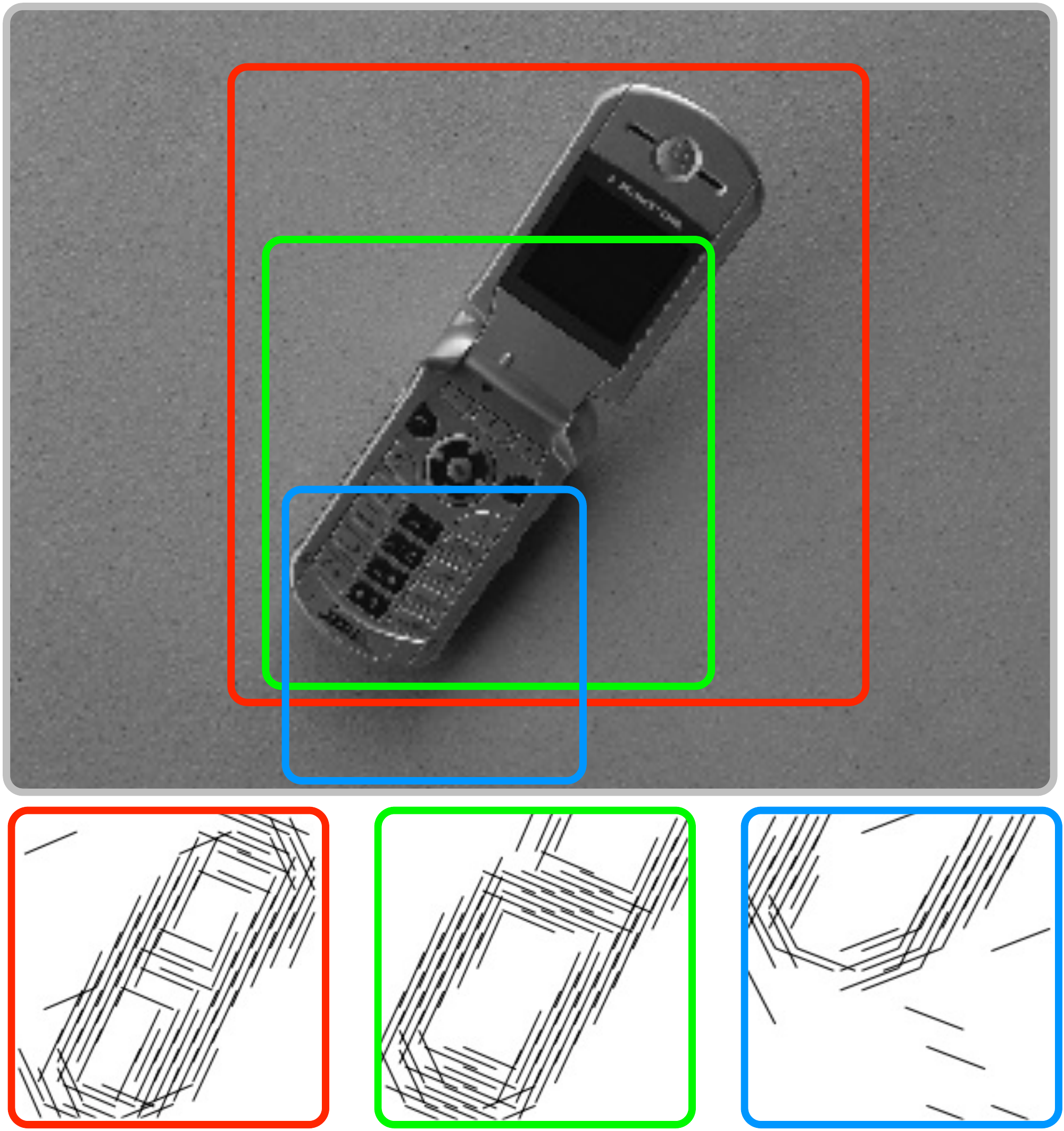}%
\label{fig_3rd_case}}
\caption{Three $S_2$ feature prototypes selective to the a) bicycle, b)face, and c) cellphone classes of 3D-object dataset along with their reconstructed preferred stimuli. It can be seen that the features converged to specific and salient object parts and neglected the irrelevant backgrounds.}
\label{features}
\end{figure*}
\begin{figure*}

\centering
\subfloat[View 1.]{\includegraphics[width=1.7in]{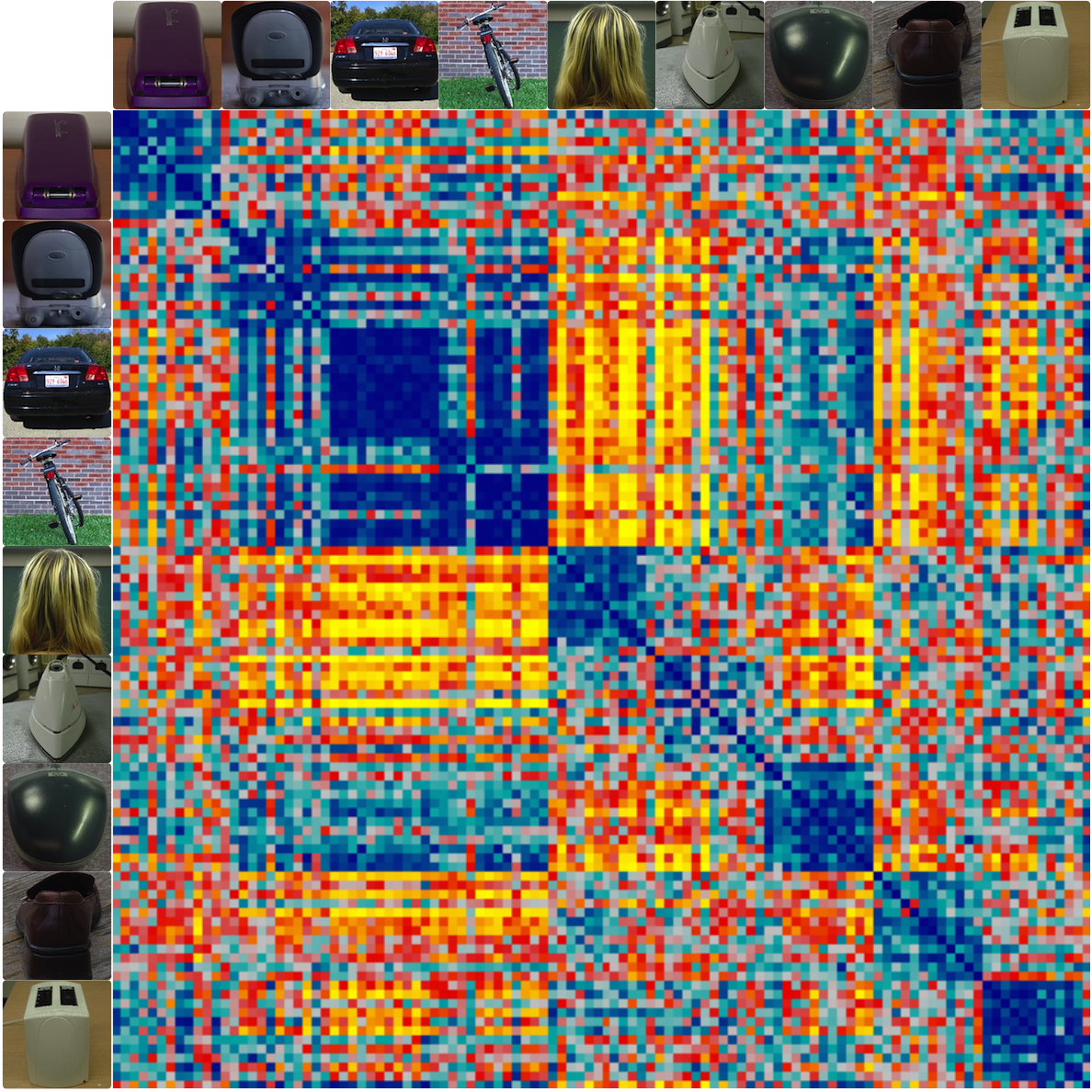}%
\label{fig_first_case2}}
\hfil
\subfloat[View 2.]{\includegraphics[width=1.7in]{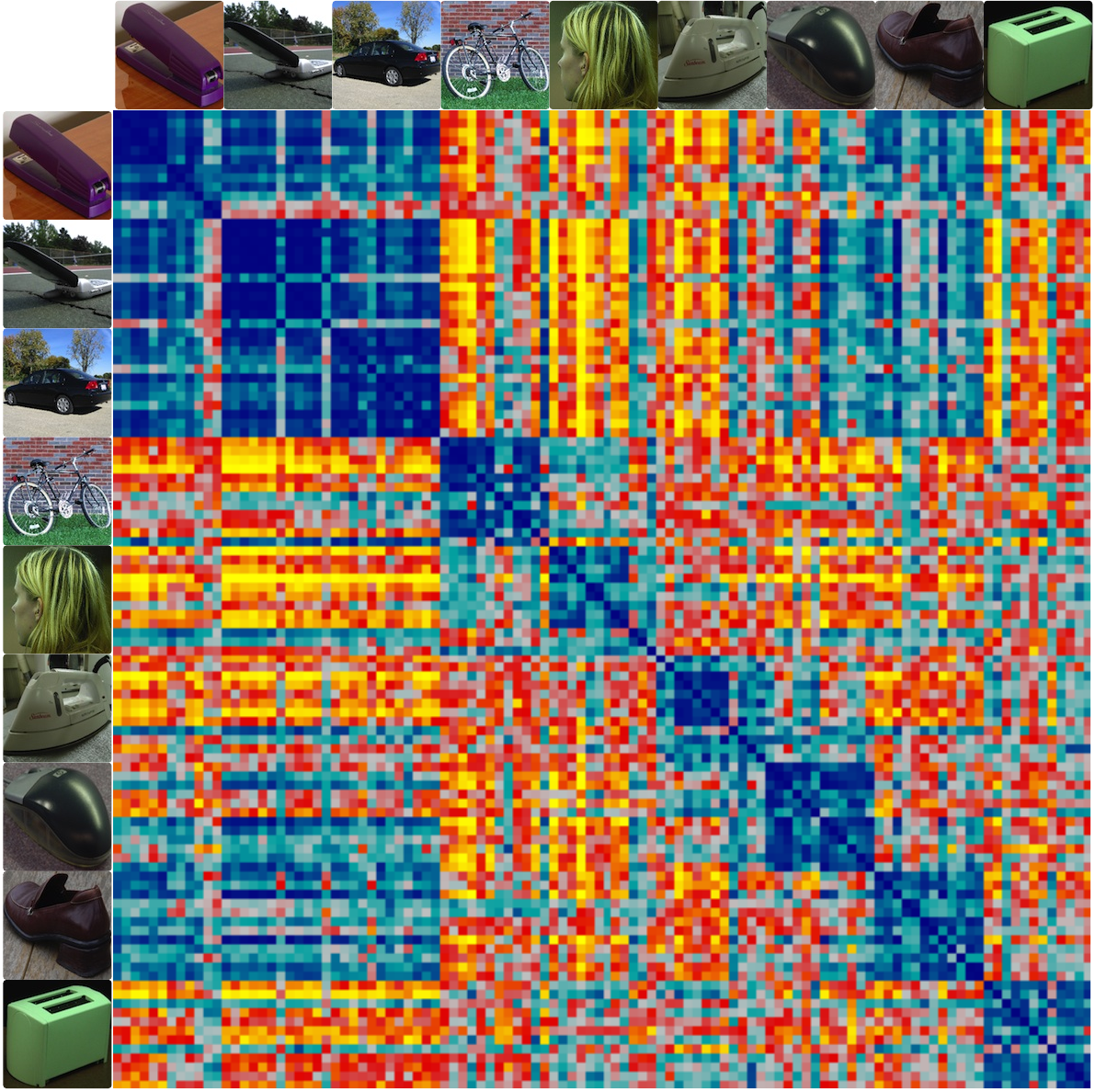}%
\label{fig_second_case2}}
\hfil
\subfloat[View 3.]{\includegraphics[width=1.7in]{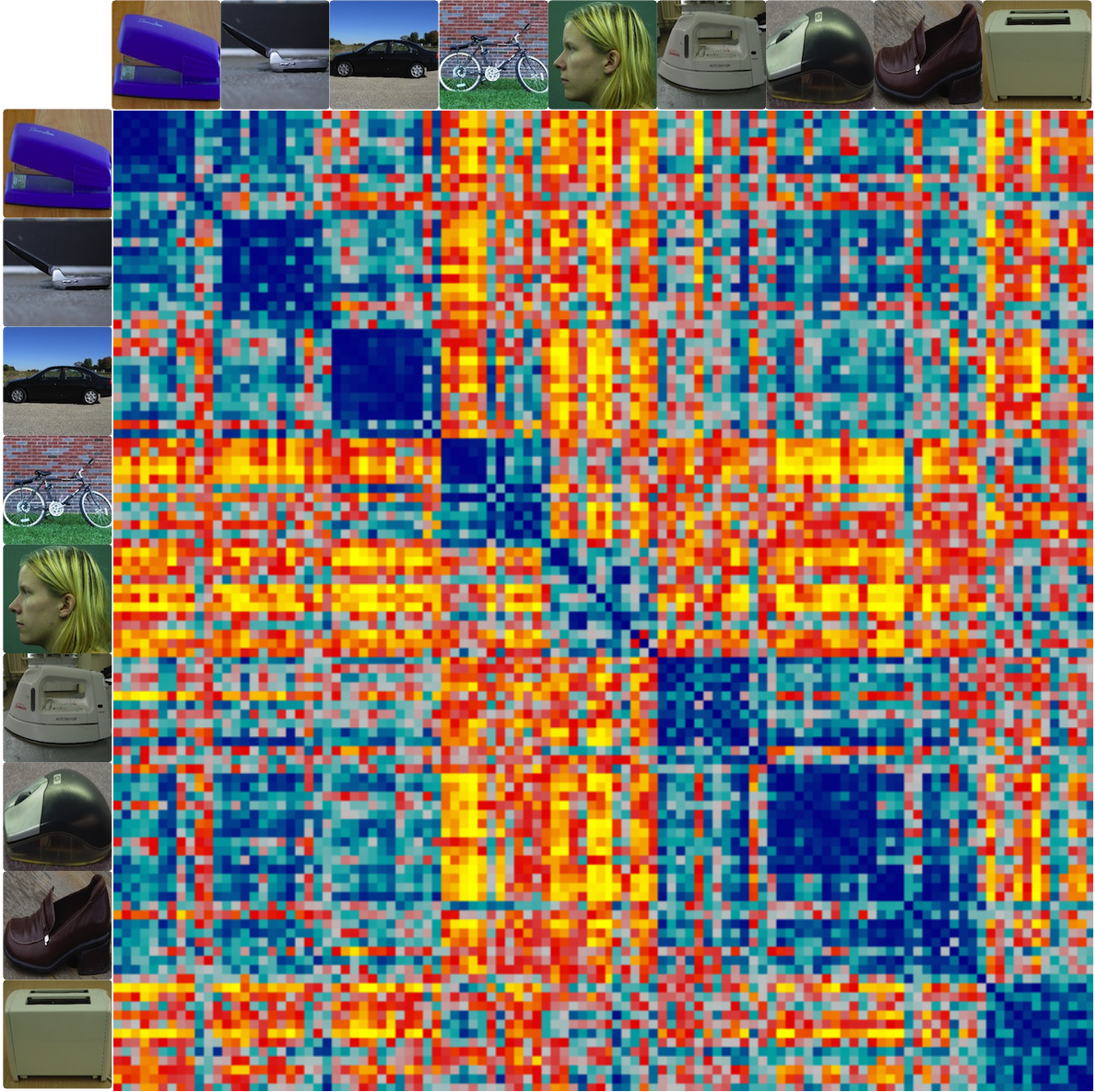}%
\label{fig_3rd_case3}}
\hfil
\subfloat[View 4.]{\includegraphics[width=1.7in]{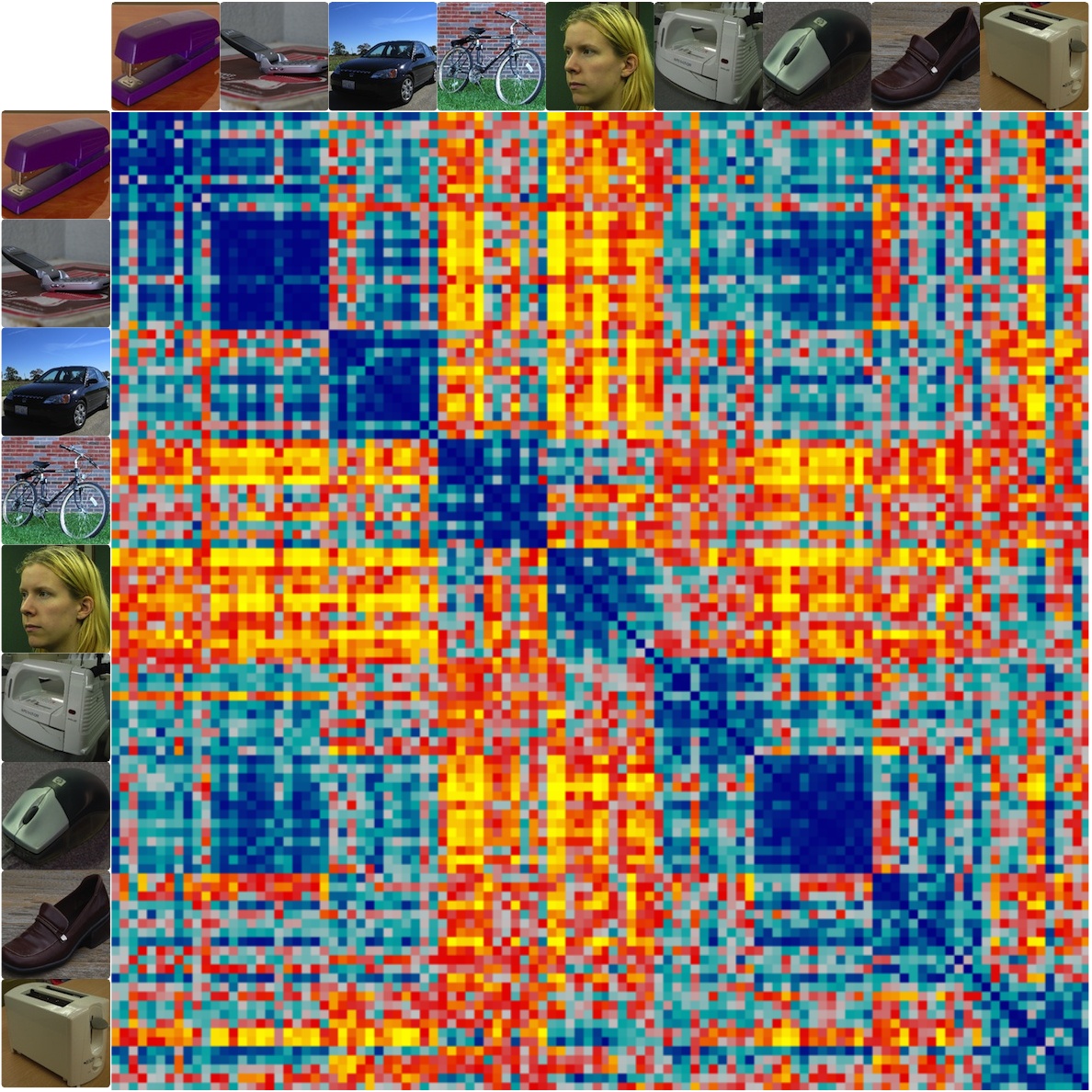}%
\label{fig_4th_case2}}
\hfil
\subfloat[View 5.]{\includegraphics[width=1.7in]{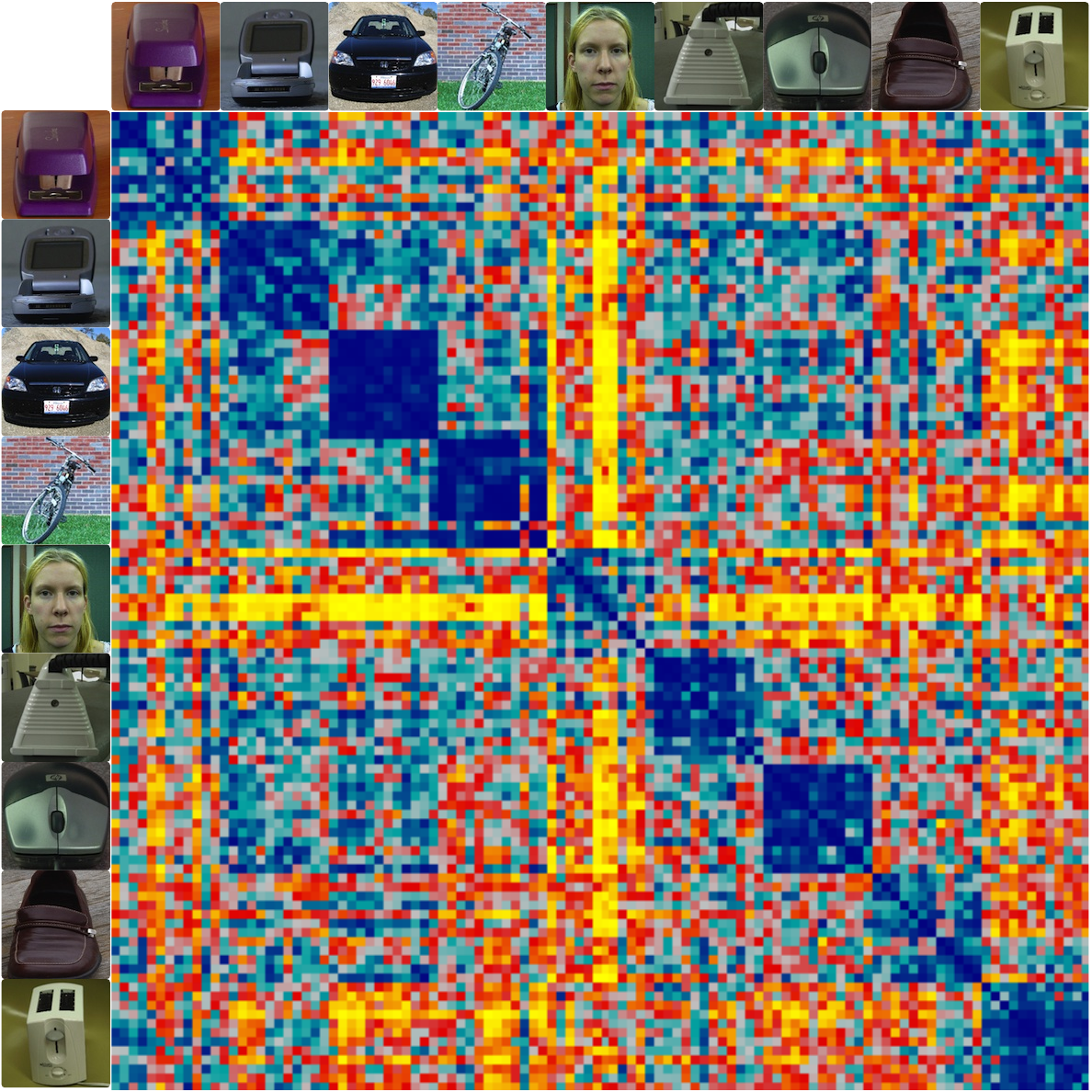}%
\label{fig_5th_case2}}
\hfil
\subfloat[View 6.]{\includegraphics[width=1.7in]{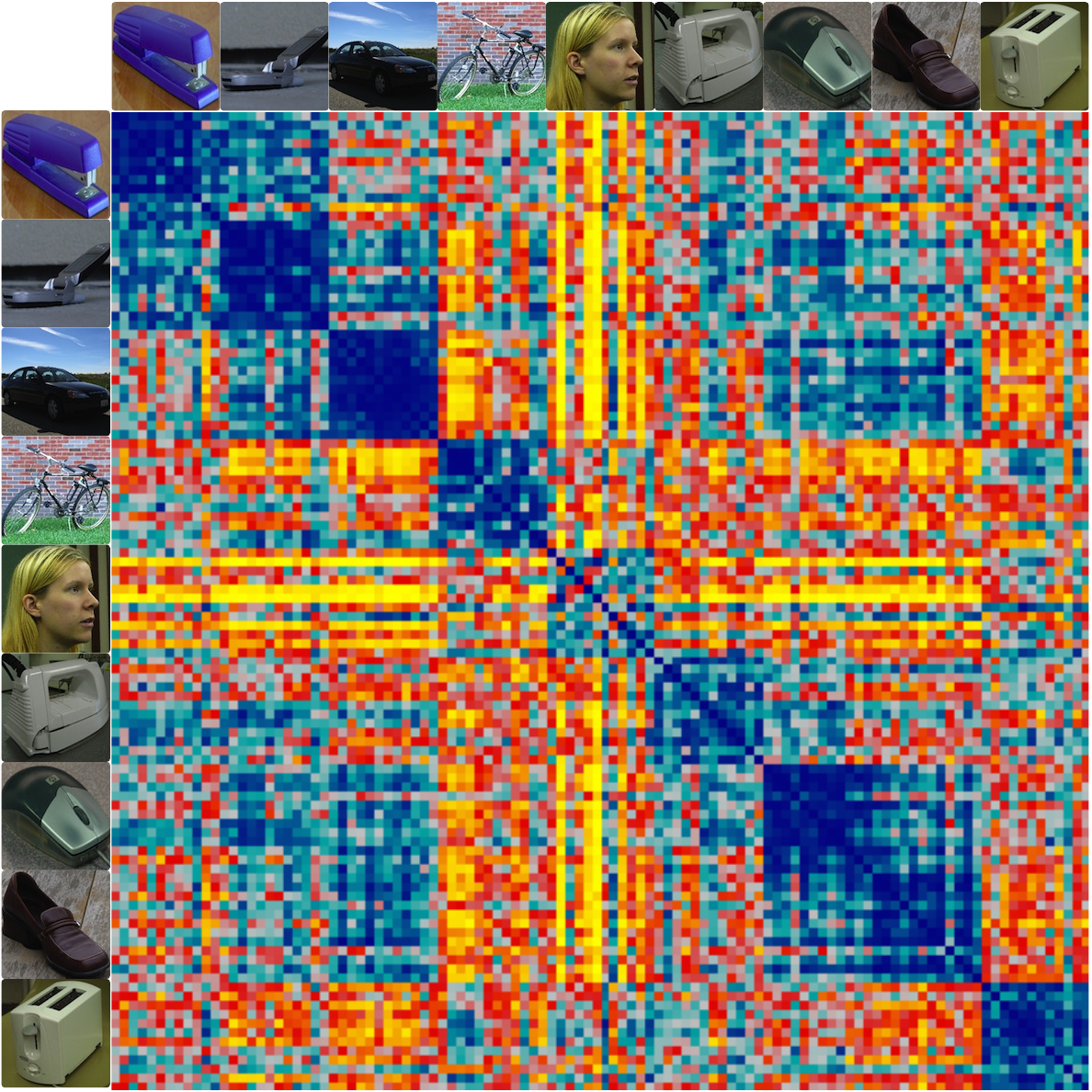}%
\label{fig_6th_case2}}
\hfil
\subfloat[View 7.]{\includegraphics[width=1.7in]{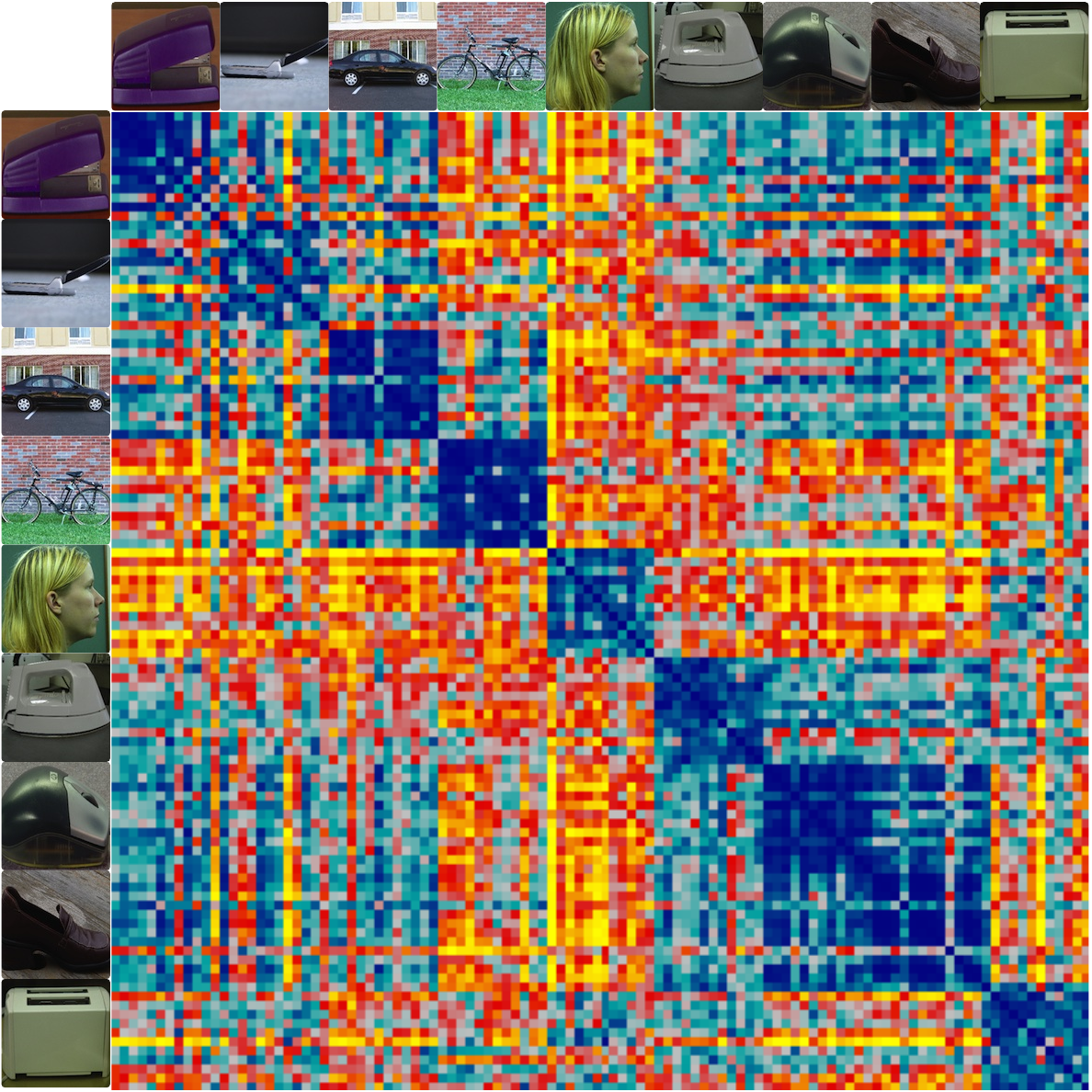}%
\label{fig_7th_case2}}
\hfil
\subfloat[View 8.]{\includegraphics[width=1.7in]{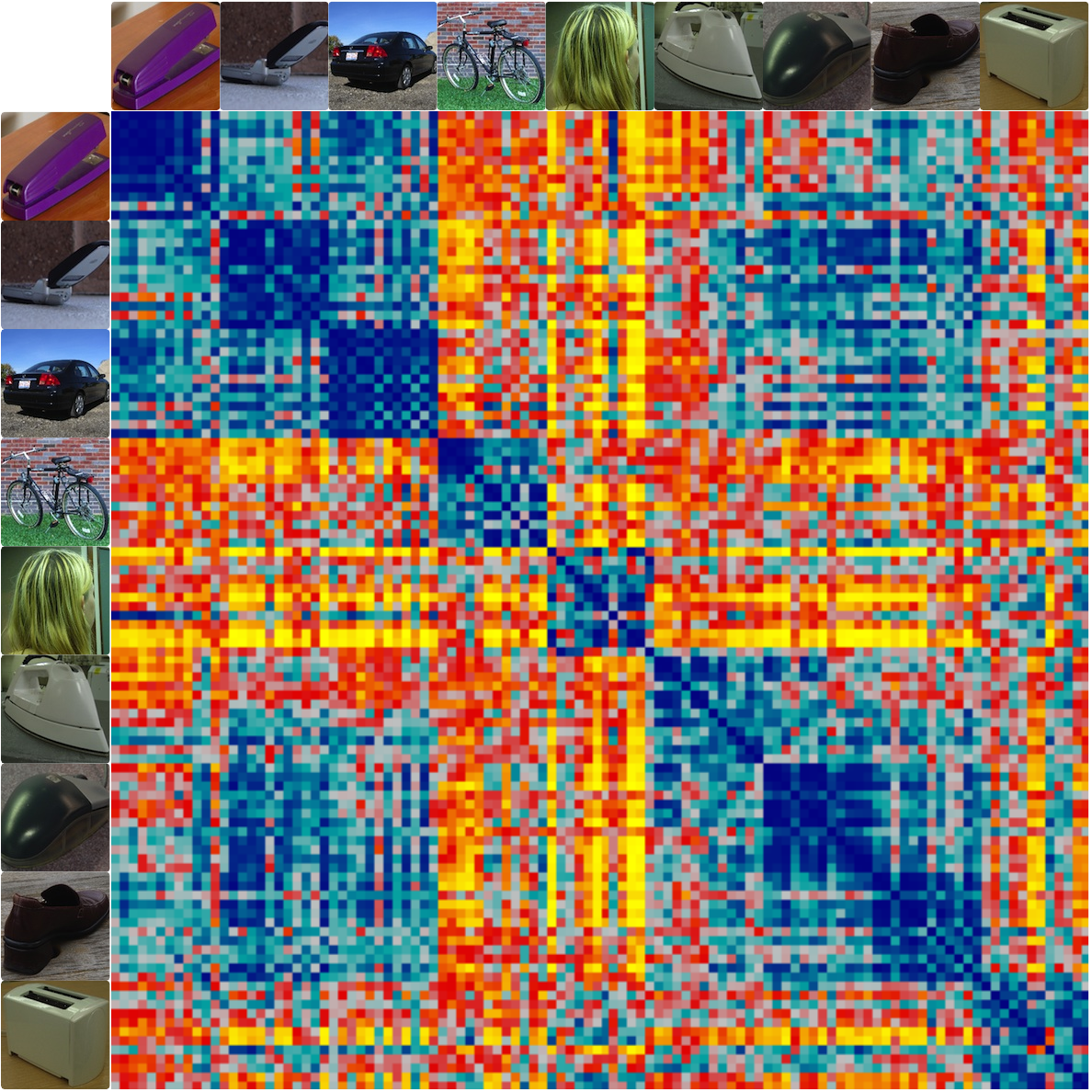}%
\label{fig_8th_case2}}
\hfil
\vspace*{0.5cm}
\captionsetup[subfigure]{labelformat=empty}
\subfloat[][Dissimilarity measure]{\includegraphics[width=2.2in]{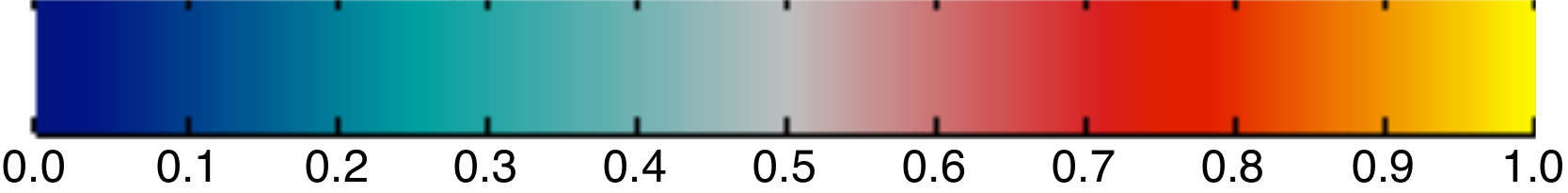}}%

\caption{RDMs of our model on 3D-Object dataset corresponding to different viewpoints. It can be seen that within class dissimilarities are very low (the blue squares around the main diagonal where rows and columns correspond to images of the same category), while between class dissimilarities are higher (more yellowish). Note that due to the absence of image samples for some views of the monitor class, we have eliminated this class from the RDMs.}
\label{RDMs}
\end{figure*}
 \begin{figure*}[t]
\centering
\subfloat[View 1.]{\includegraphics[width=1.7in]{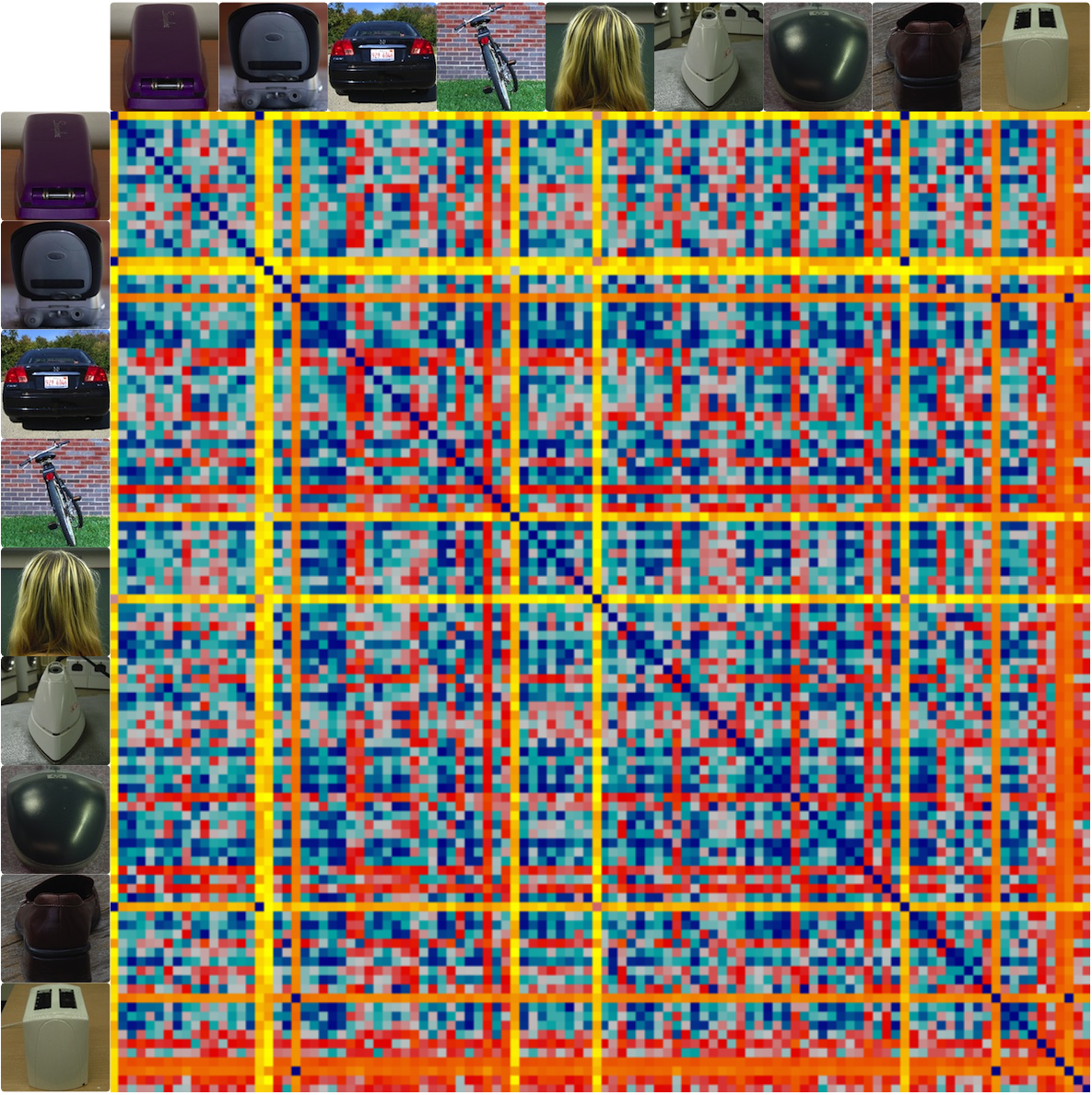}%
\label{Hmax_fig_first_case2}}
\hfil
\subfloat[View 2.]{\includegraphics[width=1.7in]{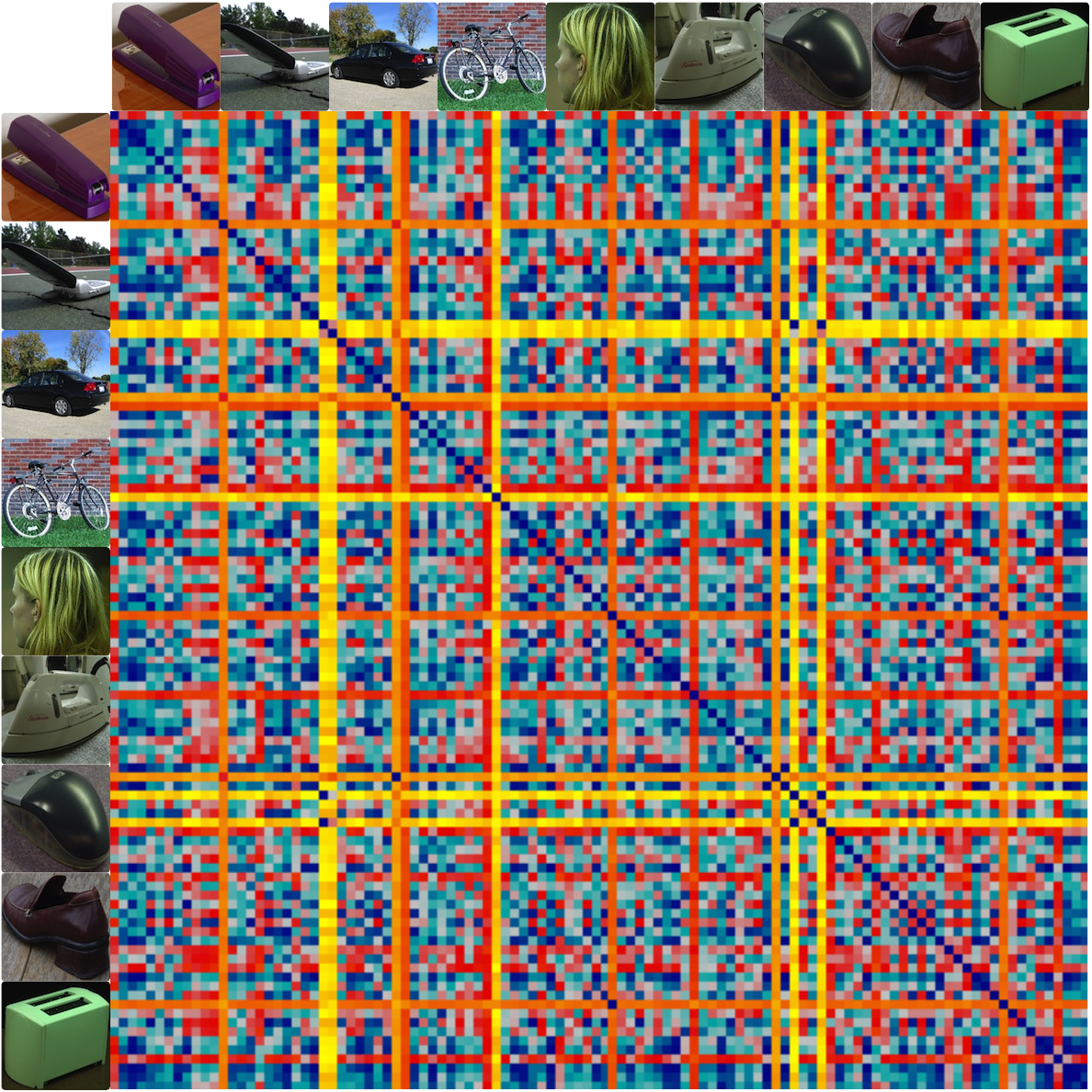}%
\label{Hmax_fig_second_case2}}
\hfil
\subfloat[View 3.]{\includegraphics[width=1.7in]{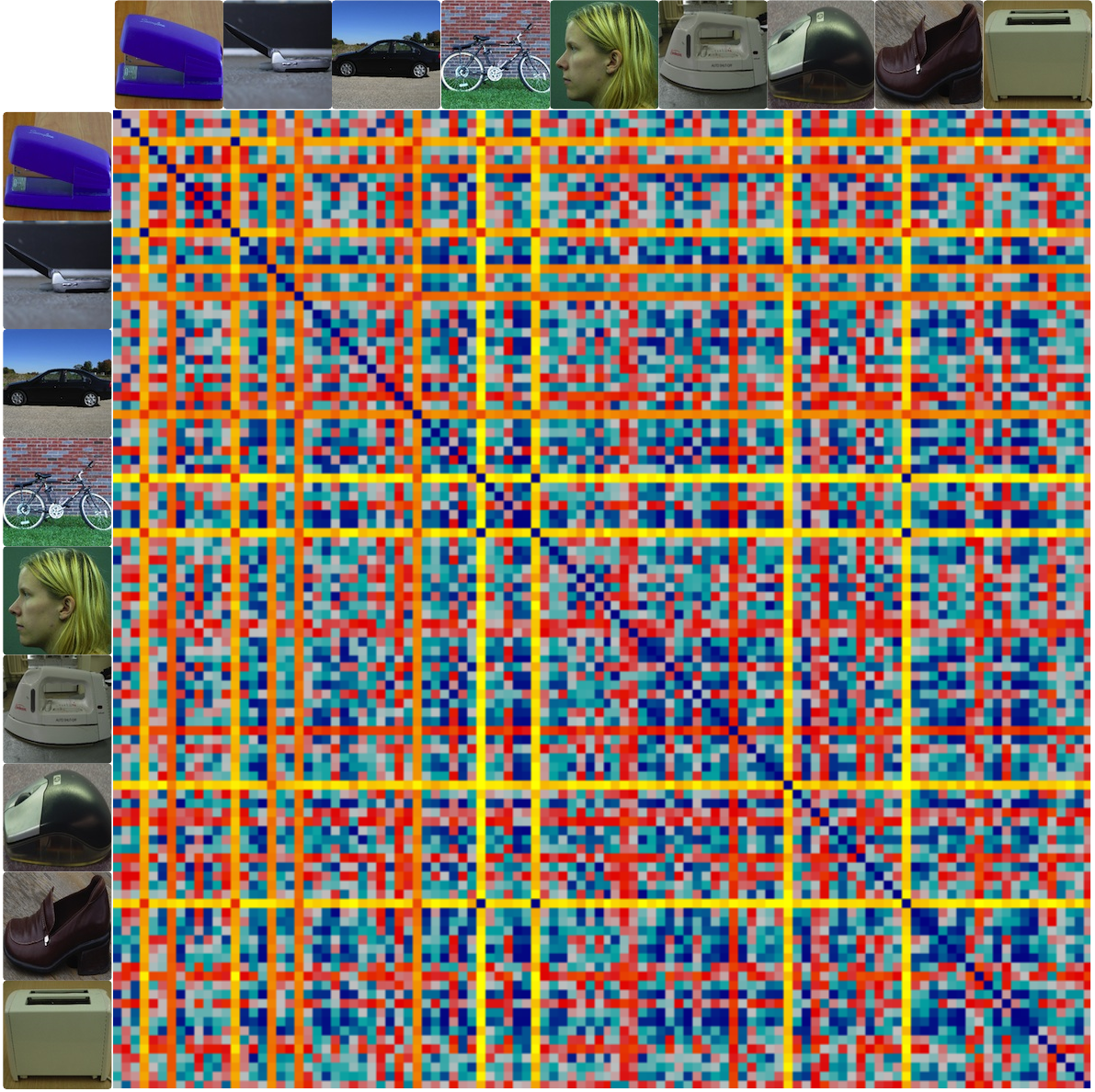}%
\label{Hmax_fig_3rd_case3}}
\hfil
\subfloat[View 4.]{\includegraphics[width=1.7in]{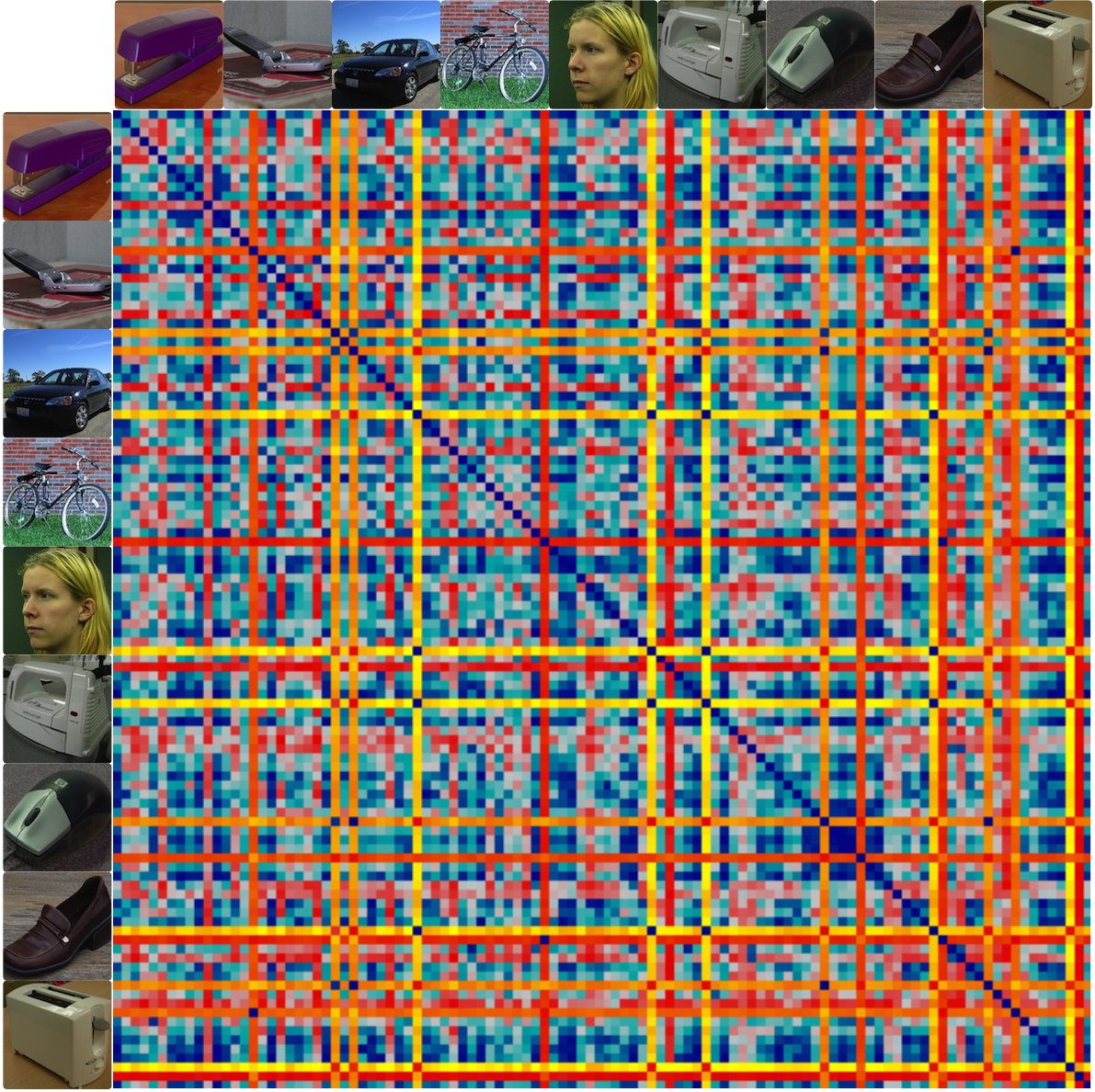}%
\label{Hmax_fig_4th_case2}}
\hfil
\subfloat[View 5.]{\includegraphics[width=1.7in]{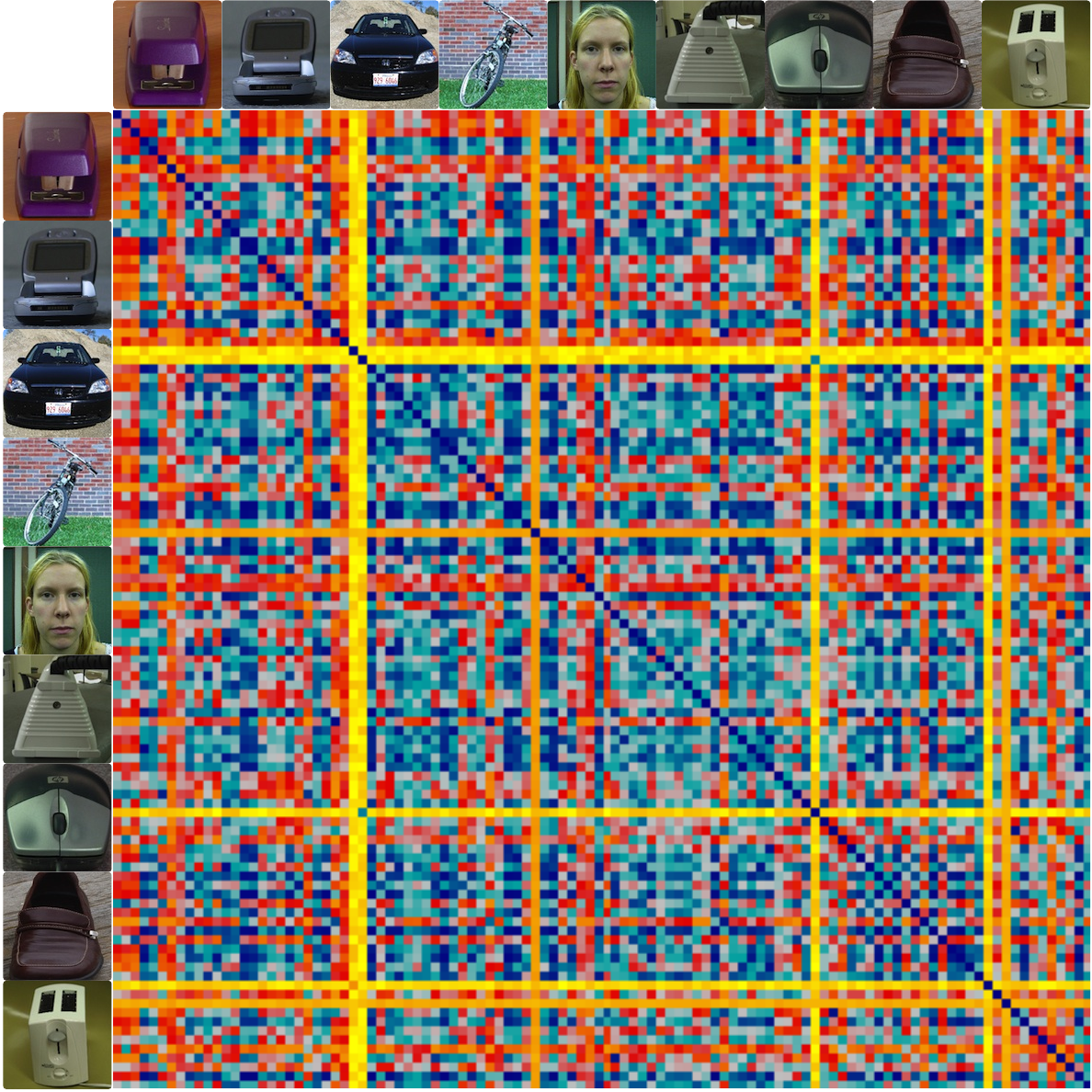}%
\label{Hmax_fig_5th_case2}}
\hfil
\subfloat[View 6.]{\includegraphics[width=1.7in]{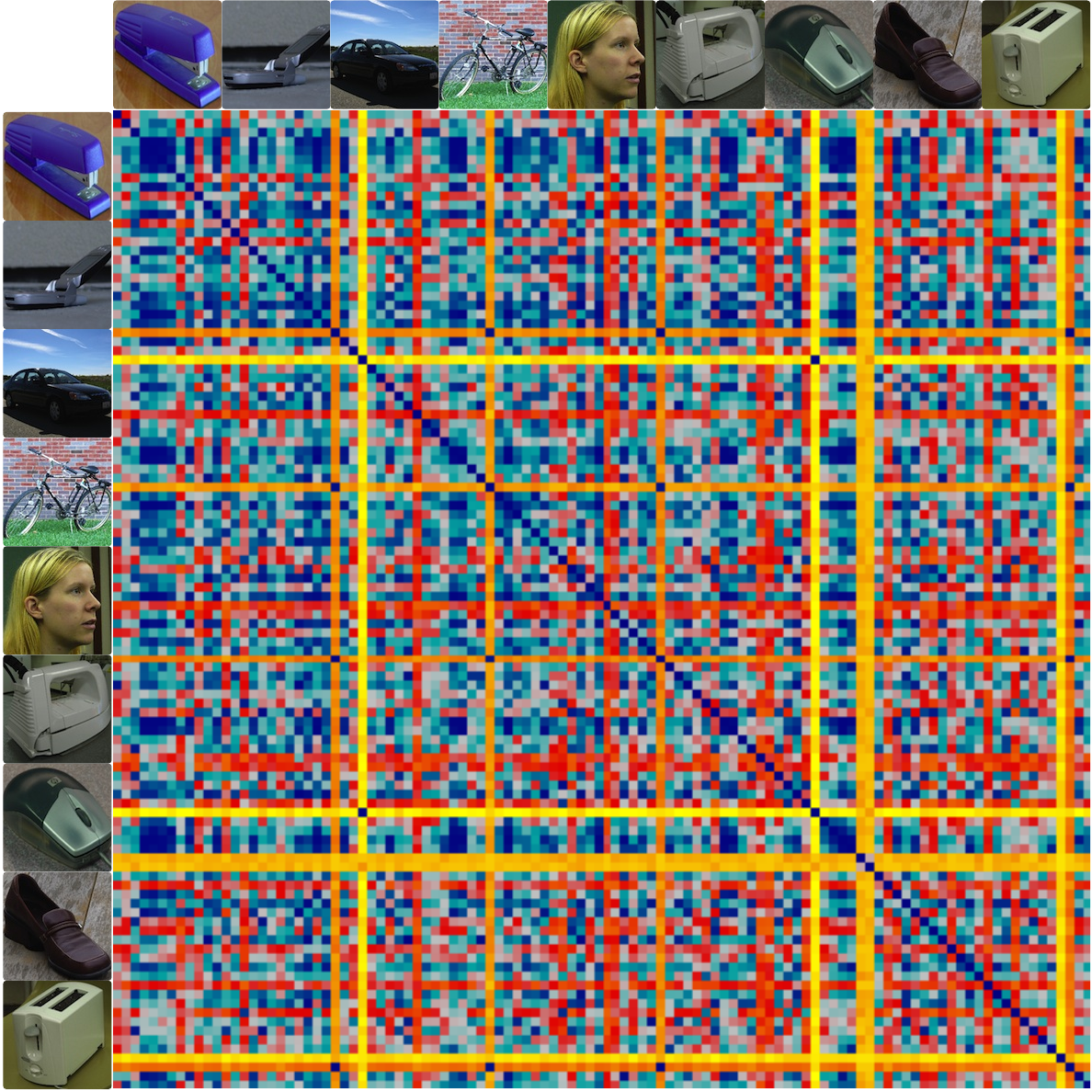}%
\label{Hmax_fig_6th_case2}}
\hfil
\subfloat[View 7.]{\includegraphics[width=1.7in]{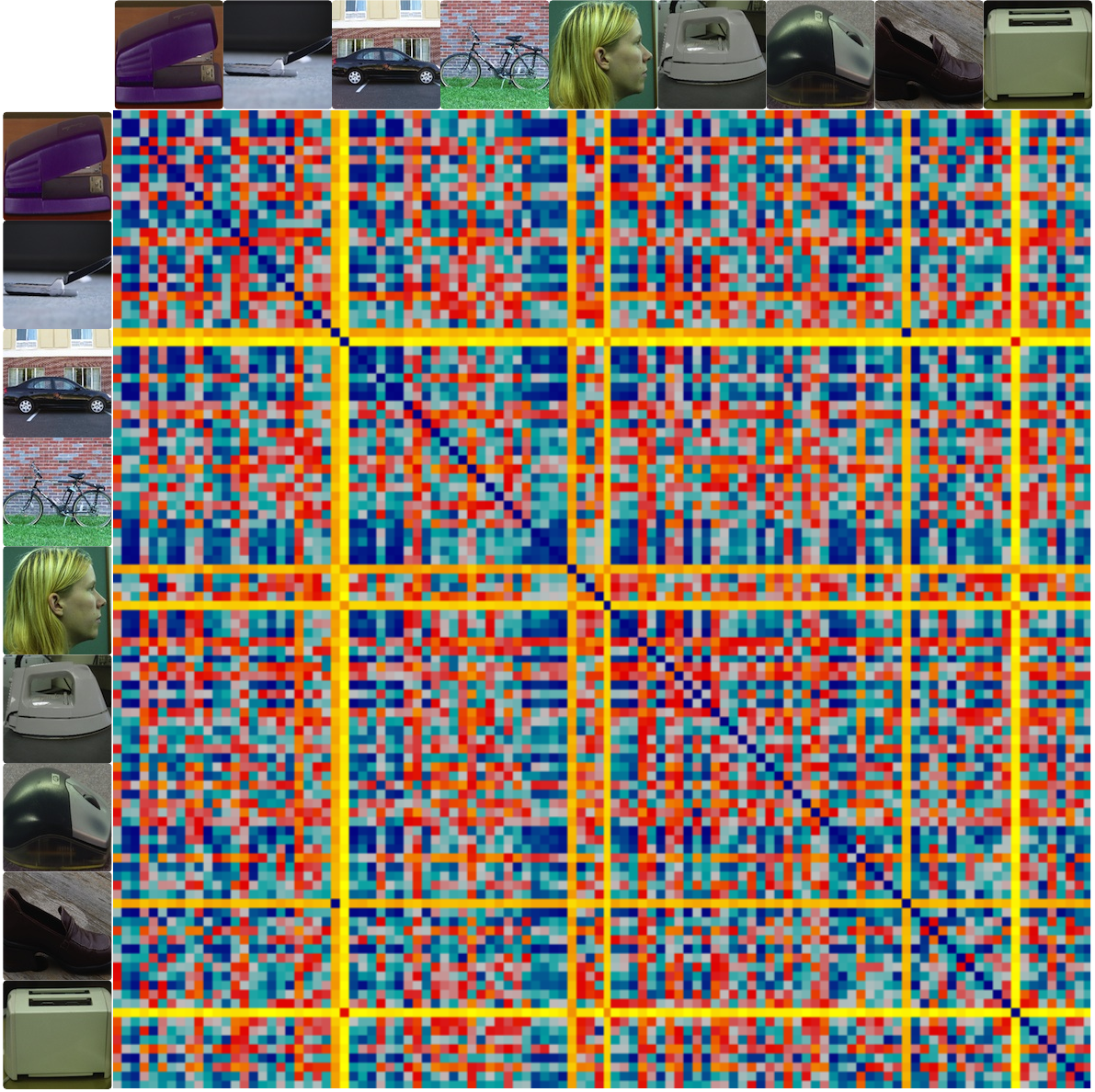}%
\label{Hmax_fig_7th_case2}}
\hfil
\subfloat[View 8.]{\includegraphics[width=1.7in]{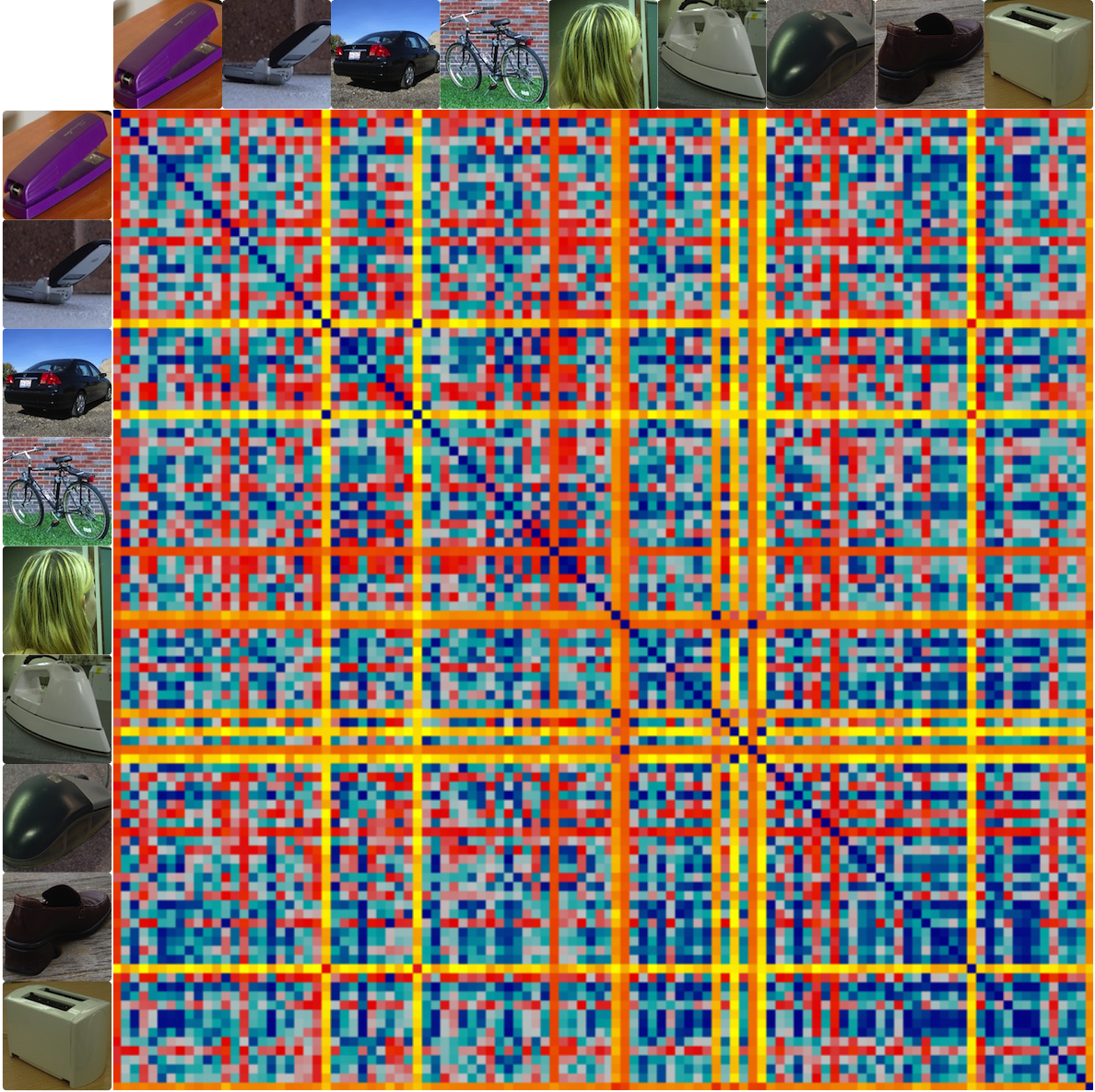}%
\label{Hmax_fig_8th_case2}}
\hfil
\vspace*{0.5cm}
\captionsetup[subfigure]{labelformat=empty}
\subfloat[][Dissimilarity measure]{\includegraphics[width=2.2in]{Legend}}%

\caption{RDMs of the HMAX model on 3D-Object dataset corresponding to different viewpoints. Randomly selected features in HMAX  model are not able to similarly represent within-category objects and dissimilarly represent between-category objects. Note that due to the absence of image samples for some views of the monitor class, we have eliminated this class from the RDMs.}
\label{Hmax_RDMs}
\end{figure*} 

In this section, we demonstrate that unsupervised STDP learning algorithm extracts informative and diagnostic features by comparing them to the randomly picked HMAX features. To this end, we have used several feature analysis techniques: representational dissimilarity matrices, hierarchical clustering, and mutual information. We performed the same analyses on both datasets and obtained similar results. Hence, the results of ETH-80 are presented in Supplementary Information.

Extraction of diagnostic features let our model reach high classification performances with a small number of features (c.f. Table~\ref{table_example1}). To understand why this is true, we first reconstructed the features' preferred stimuli. Given that each $S_2$ neuron receives spikes from $C_1$ neurons responding to bars in different orientations, the representation of the preferred features of $S_2$ neurons could be reconstructed by convolving their weight matrices with a set of kernels representing oriented bars. In Fig.~\ref{features} the receptive fields of activated $S_2$ neurons along with the representation of their preferred stimuli are illustrated (Fig. S2 provides the same illustration for the ETH-80 dataset). This demonstrates that only a small number of $S_2$ neurons are required to represent the input objects. In other words, the obtained features are compatible with the sparse coding theory in visual cortex. In addition, for an input image, the most activated $S_2$ neurons cover the input objects and they do not respond to the background area. Indeed, the STDP learning algorithm naturally focuses on what are common in the training images, which are the target object features. The backgrounds are generally not learned (at least not in priority), since they are almost always too different from one image to another and the STDP process cannot converge on them.

To characterize the neuronal population coding in the $C_2$ layer of the model and to study the quality of  $C_2$ features, we used the representational dissimilarity matrix (RDM)~\cite{kriegeskorte2008representational}). Each element of the RDM reflects the measure of dissimilarity (distance) among the neural activity patterns (i.e., the object representations) associated with two different image stimuli. The distance we used here is $1-Pearson\:\:correlation$. In an RDM corresponding to a perfect model, the representations of the objects of the same category have low dissimilarities (i.e., highly correlated), whereas objects of  different categories are represented highly dissimilarly (i.e., uncorrelated). Hence, if we group the rows and columns of the RDM of a perfect model based on object categories, it is expected to see squares of low dissimilarity values around the main diagonal, each of which corresponds to pairs of same-category images, while other elements have higher values.

Here, to plot the RDM of each view angle, first, the images of all input instances which are taken in that view are picked. Then the corresponding RDM is plotted by computing the pairwise dissimilarity of the values of $C_2$ features associated with each pair of images. Figure~\ref{RDMs} presents the RDMs of our model for all eight views (see Fig. S3 for ETH-80). In each RDM, rows and columns are sorted based on image categories. Also, a sample image of each category is placed next to the rows and columns which correspond to that category. Here, we used a color-code to represent RDMs which ranges from pure blue to pure yellow demonstrating low to high dissimilarities, respectively. It can be seen that the within-category dissimilarity values (identified by blue squares around the main diagonal) are relatively lower than the between-category dissimilarities (more yellowish areas). As expected, the RDMs indicate that the obtained performance is not due to the capabilities of the classifier, but to the extraction of diagnostic and highly informative $C_2$ features through STDP.

We have also computed the RDMs of the HMAX model (including 12000 features) for  eight views, as provided in Fig.~\ref{Hmax_RDMs} (see Fig. S4 for ETH-80). As it can be seen in this figure, the randomly selected features in the HMAX model are unable to similarly represent within-category objects and dissimilarly represent between-category objects. This is probably due to uninformative features used by the HMAX model. Indeed, in HMAX, the task of selecting the informative features is left to the classifier. We also note the presence of horizontal and vertical yellow lines, indicating ``outliers", whose representation lies far away from all the others. This indicates that the features do not pave well the stimulus space.
\begin{figure*}[!htb]
\centering
\includegraphics[scale=0.75]{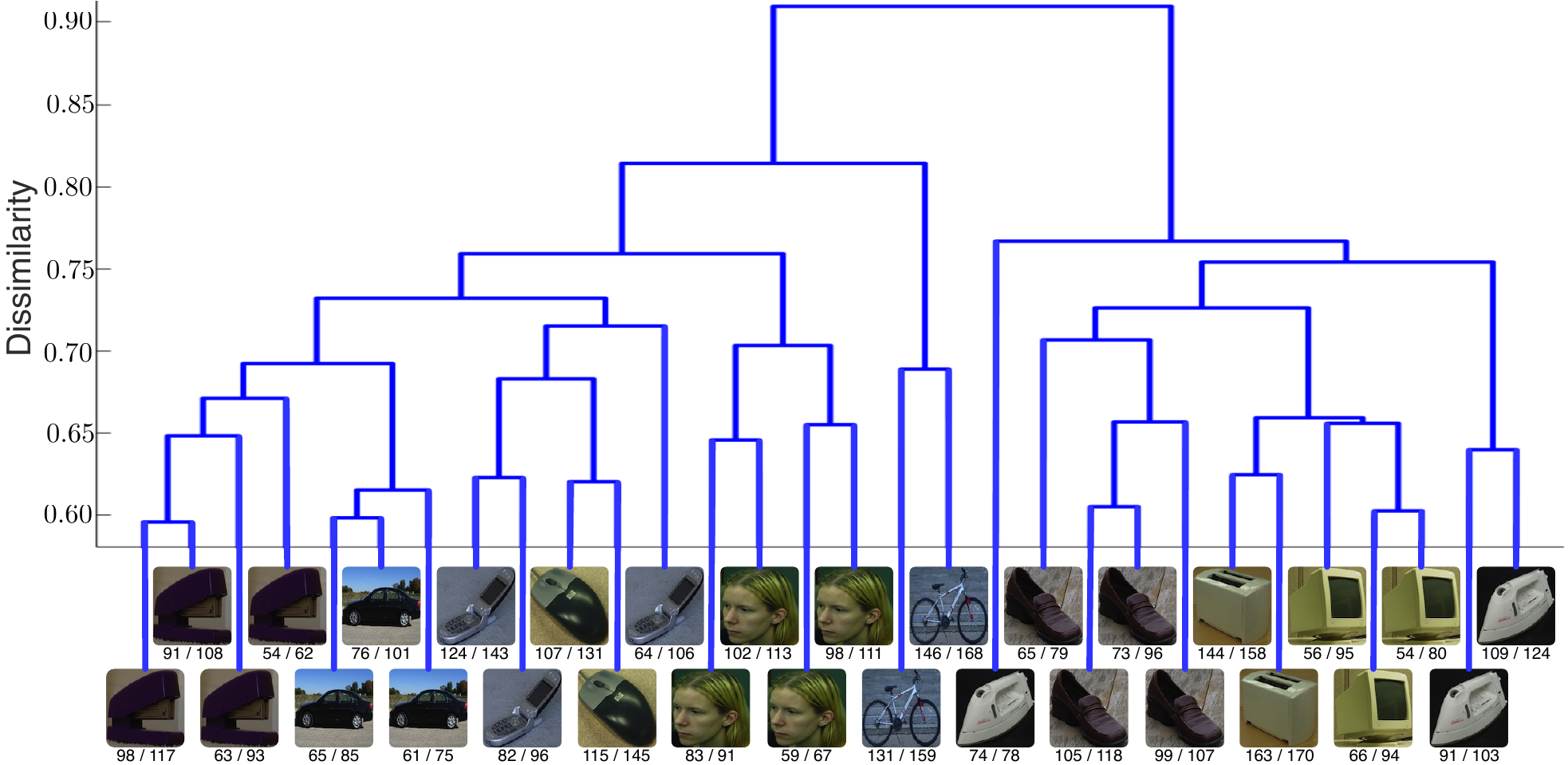}
\caption{The hierarchy of clusters and their labels for our model on 3D-Object dataset. The label of each cluster indicates the class with the highest frequency in that cluster. It can be seen that the samples of each class are placed in close clusters. The cardinality of each cluster, $C$, and the cardinality of the class with the highest frequency, $H$, are placed below the cluster label as $H/C$.}
\label{Hierarchy}
\end{figure*} 
\begin{figure*}[!htb]
\centering
\includegraphics[scale=0.75]{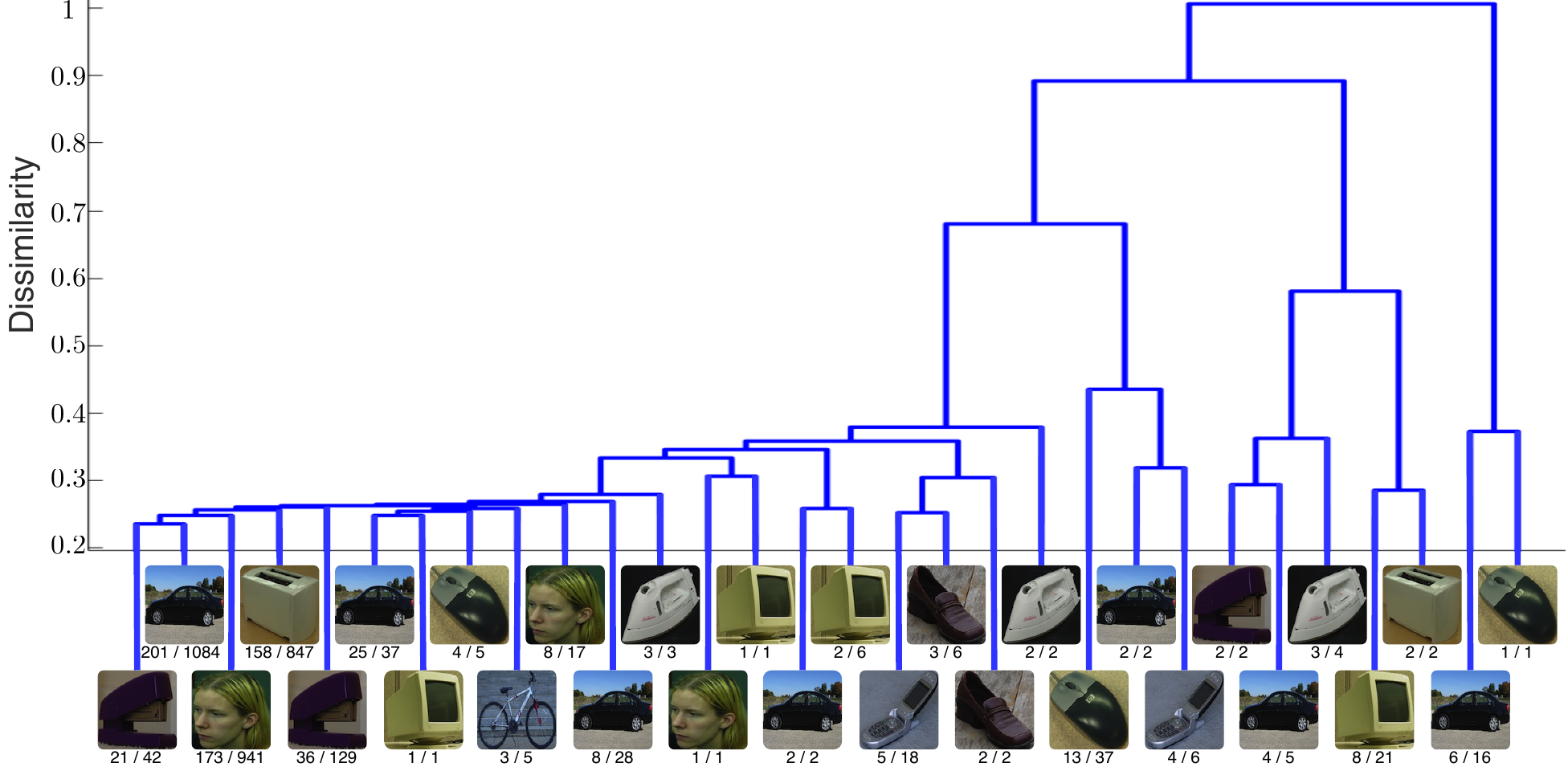}
\caption{The hierarchy of clusters and their labels for the HMAX model on 3D-Object dataset. The label of each cluster indicates the class with the highest frequency in that cluster. It can be seen that the majority of the objects are assigned to a small number of clusters and samples of each class are not well placed in close clusters. The cardinality of each cluster, $C$, and the cardinality of the class with the highest frequency, $H$, are placed below the cluster label as $H/C$.}
\label{Cluster5_HMax}
\end{figure*} 

To see how well the stimuli are distributed in the high dimensional feature space, we performed hierarchical clustering over the test set. The clustering procedure is started by considering each stimulus as a discrete cluster node, continued by connecting the closest nodes into a new combined cluster node, and completed by connecting all the stimuli to a single node. We performed this analysis on the $C_2$ feature vectors corresponding to all objects in all views, scales, and tilts. The obtained hierarchy for our model is displayed in Fig.~\ref{Hierarchy} (see Fig. S5 for ETH-80). The distance between a pair of cluster nodes is computed by measuring the dissimilarity among their centers (the average of cluster members). Due to the large number of stimuli, it is not possible to plot the whole hierarchy, hence, only the high level clusters are shown in this figure. For each lowest level cluster, the class with the highest frequency is illustrated by an image label. The cardinality of this class as well as the cardinality of the cluster are shown below the labels. It can be seen that the instances of each object class are placed in neighboring regions of the $C_2$ feature space. By considering the obtained hierarchical clustering and the classification accuracies, it can be concluded that the $C_2$ features are able to invariantly represent the objects in such a way that the classifier can easily separate them. 

The same hierarchical clustering is performed for the HMAX feature space (with 12000 features), as depicted in Fig.~\ref{Cluster5_HMax} (see Fig. S6 for ETH-80). As it can be seen, the majority of clusters are small, and contrary to our model, the distances between the clusters are very low. In other words, the objects are densely represented in a small area of such a high dimensional feature space. Furthermore, the mean intra- and inter-class dissimilarities in our model are equal to 0.40 and 0.70, respectively, while these statistics for the HMAX model are equal to 0.27 and 0.29, respectively. In summary, it can be concluded that the distribution of the object classes are dense and highly overlapped in the HMAX feature space, while the object classes are well separated in the feature space of our model. 

\begin{table*}
\caption{The number of common features between each pair of classes of 3D-Object dataset. The gray level of each cell indicates the relative distance of the cell value to the maximum possible value (=50).}
\label{table_example2}
\centering
\begin{tabular}{|c|c|c|c|c|c|c|c|c|c|c|}
\hline
Class & Bicycle & Car & Cellphone & Head & Iron & Monitor & Mouse & Shoe & Stapler & Toaster\\
\hline
Bicycle & \cellcolor[gray]{0.3} 50 & \cellcolor[gray]{1}0 & \cellcolor[gray]{0.98}1 & \cellcolor[gray]{0.95}3 & \cellcolor[gray]{0.91}6 &\cellcolor[gray]{0.94} 4 & \cellcolor[gray]{0.83}12 & \cellcolor[gray]{0.91}6 & \cellcolor[gray]{0.925}5 &\cellcolor[gray]{0.925} 5\\
\hline
Car & \cellcolor[gray]{1}0 & \cellcolor[gray]{0.3} 50 & \cellcolor[gray]{0.98}1 & \cellcolor[gray]{0.805}14 & \cellcolor[gray]{0.86}10 & \cellcolor[gray]{0.87}11 & \cellcolor[gray]{0.98} 1 &\cellcolor[gray]{0.96} 2 &\cellcolor[gray]{0.91} 6 &\cellcolor[gray]{0.895} 7\\
\hline
Cellphone &\cellcolor[gray]{0.98} 1 &\cellcolor[gray]{0.98} 1 & \cellcolor[gray]{0.3} 50 &\cellcolor[gray]{1} 0 &\cellcolor[gray]{0.95} 3 &\cellcolor[gray]{0.94} 4 &\cellcolor[gray]{0.96} 2 &\cellcolor[gray]{1} 0 & \cellcolor[gray]{0.86}10 &\cellcolor[gray]{0.88} 9\\
\hline
Head & \cellcolor[gray]{0.95}3 &\cellcolor[gray]{0.805} 14 &\cellcolor[gray]{1} 0 & \cellcolor[gray]{0.3} 50 & \cellcolor[gray]{1}0 & \cellcolor[gray]{1}0 &\cellcolor[gray]{0.86} 10 & \cellcolor[gray]{0.775}16 &\cellcolor[gray]{0.96} 2 &\cellcolor[gray]{0.96} 2\\
\hline
Iron & \cellcolor[gray]{0.91}6 & \cellcolor[gray]{0.86}10 &\cellcolor[gray]{0.95} 3 &\cellcolor[gray]{1} 0 & \cellcolor[gray]{0.3} 50 & \cellcolor[gray]{0.70}21 & \cellcolor[gray]{1}0 &\cellcolor[gray]{0.94} 4 &\cellcolor[gray]{0.83} 12 & \cellcolor[gray]{0.91}6\\
\hline
Monitor &\cellcolor[gray]{0.94} 4 &\cellcolor[gray]{0.87} 11 &\cellcolor[gray]{0.94} 4 &\cellcolor[gray]{1} 0 & \cellcolor[gray]{0.70}21 & \cellcolor[gray]{0.3} 50 & \cellcolor[gray]{1}0 &\cellcolor[gray]{1} 0 &\cellcolor[gray]{0.88} 8 & \cellcolor[gray]{0.805}14\\
\hline
Mouse & \cellcolor[gray]{0.83}12 &\cellcolor[gray]{0.98} 1 &\cellcolor[gray]{0.96} 2 & \cellcolor[gray]{0.86}10 & 0 & 0 & \cellcolor[gray]{0.3} 50 & \cellcolor[gray]{0.96}2 & \cellcolor[gray]{0.96}2 & \cellcolor[gray]{0.925}5\\
\hline
Shoe & \cellcolor[gray]{0.91}6 & \cellcolor[gray]{0.96}2 & \cellcolor[gray]{1}0 & \cellcolor[gray]{0.775}16 & \cellcolor[gray]{0.94}4 & \cellcolor[gray]{1}0 & \cellcolor[gray]{0.96}2 & \cellcolor[gray]{0.3} 50 & \cellcolor[gray]{1}0 & \cellcolor[gray]{1}0\\
\hline
Stapler & \cellcolor[gray]{0.925}5 & \cellcolor[gray]{0.91}6 & \cellcolor[gray]{0.86}10 & \cellcolor[gray]{0.96}2 & \cellcolor[gray]{0.83}12 & \cellcolor[gray]{0.88}8 &\cellcolor[gray]{0.96} 2 &\cellcolor[gray]{1} 0 & \cellcolor[gray]{0.3} 50 & \cellcolor[gray]{0.72}20\\
\hline
Toaster & \cellcolor[gray]{0.925}5 & \cellcolor[gray]{0.895}7 & \cellcolor[gray]{0.88}9 & \cellcolor[gray]{0.96}2 & \cellcolor[gray]{0.91}6 & \cellcolor[gray]{0.805}14 & \cellcolor[gray]{0.925}5 & \cellcolor[gray]{1}0 & \cellcolor[gray]{0.72}20 & \cellcolor[gray]{0.3} 50\\
\hline
\end{tabular}
\end{table*}

In an other experiment, we analyzed the class dependency of the $C_2$ features for our model. To this end, the 50 most informative features, when classifying a specific class against all the other classes, are selected by employing the mutual information technique. In other words, for each class, we selected those 50 features which have the highest activity for samples of that class and have less activity for other classes. Afterwards, the number of common features among the informative features of each pair of classes are computed as provided in Table~\ref{table_example2}. On average, there are only about 5.4 common features between pairs of classes. Although there are some common features between any two classes,  their co-occurrence with the other features help the classifier to separate them from each other. In this way, our model can represent various object classes with a relatively small number of features. Indeed, exploiting the intermediate complexity features, which are not common in all classes and are not very rare, can help the classifier to discriminate instances of different classes~\cite{Ullman2002}.

\subsection{Random features and simple classifier}
In a previous study~\cite{leibo2010primal}, it has been shown that using the HMAX model with random dot patterns in the $S_{2}$ layer can reach  a reasonable performance, comparable to the one obtained with random  patches cropped from the training images. It seems that this is due to the dependency of HMAX to the application of a powerful classifier. Indeed, the use of both random dot or randomly selected patches transform the images into a complex and nested feature space and it is the classifier which looks for a complex signature to separate object classes. The deficiencies emerge when the classification problem gets harder (such as invariant or multiclass object recognition problems) and then even a powerful classifier is not able to discriminate the classes~\cite{Pinto2008,pinto2011comparing}. Here, we show that the superiority of our model is due to the  informative feature extraction through a bio-inspired learning rule. To this end,  we have  compared the performances on 3D-Object dataset obtained with random features versus STDP features, as well as a very simple classifier versus SVM.

To generate random features, we have set the weight matrix of each $S_{2}$ feature of our model with random values. First, we have computed the mean and standard deviation (STD) ($253 \pm 21$) of the number of active (nonzero) weights in the features learned by STDP. Second, for each random feature, the number of active weights, $N$, is computed by generating a random number based on the obtained mean and STD. Finally, a random feature is constructed by uniformly distributing the $N$ randomly generated values in the weight matrix.

In addition, we designed a  simple classifier comprised of several one-versus-one classifiers. For each binary classifier, two subset of $C_{2}$ features with high occurrence probabilities in one of the two classes are selected. In more details, to select suitable features for the first class, the occurrence probabilities of $C_{2}$ features in this class are divided by the corresponding occurrence probabilities in the second class. Then, a feature is selected if this ratio is higher than a threshold. The optimum threshold value is computed by a trial and error search in which the performance over the training samples is maximized. To assign a class label to the input test sample, we performed an inner product on the feature value and feature probability vectors. Finally, the class with the highest probability is reported to the combined classifier. The combined classifier selects the winner class based on a simple majority voting. 

For 500 random features, using the SVM and the simple classifier, our model reached classification performances of 71\% and 21\% on average, respectively. Whereas, for the learned $S_{2}$ features, both the SVM and simple classifiers attained reasonable performances of 96\% and 79\%, respectively. Based on these results, it can be concluded that the features obtained through the bio-inspired unsupervised learning projects the objects into an easily separable space, while the feature extraction by selection of random patches (drawn from the training images) or by generation of random patterns leads to a complex object representation.

\section{Discussion}\label{Discussion}
Position and scale invariance in our model are built-in, thanks to weight sharing and scaling process. Conversely, view-invariance must be obtained through the learning process. Here, we used all images of five object instances from each category (varied in all dimensions) to learn the $S_{2}$ visual features, while images of all other object instances of each category were used to test the network. Hence, the model was exposed to all possible variations during the learning to gain view-invariance. Moreover, near or opposite views of the same object shares some features which are suitable for invariant object recognition. For instance, consider the overall shape of a head, or close views of a bike wheel which could be a complete circle or an ellipse. Regarding the fact that STDP tends to learn more frequent features in different images, different views of an object could be invariantly represented based on more common features.

Our model appears to be the best choice when dealing with few object classes, but huge variations in view points. As pointed out in previous studies, both HMAX and DeepConvNet models could not handle these variations perfectly~\cite{Pinto2008,pinto2011comparing, ghodrati2014feedforward}.
 Conversely, our model is not appropriate to handle many classes, which requires thousands of features, like in the ImageNet contest, because its time complexity is roughly in $N^2$, where $N$ is the number of features  (briefly: since the number of firing neurons per image is limited, if the number of features is doubled, reaching convergence will take roughly twice as many images, and the processing time for each of them will be doubled as well). For example, extracting 4096 features in our model, the same number of features in DeepConvNet, would take about 67 times it took us to extract 500. However, parallel implementation of our algorithm could speed-up the computation time by several orders of magnitude~\cite{lemoine2013gpu}. Even in this case, we do not expect to outperform the DeepConvNet model on the ImageNet database, since only the shape similarities are taken into account in our model and the other cues such as color or texture are ignored.

Importantly, our algorithm has a natural tendency to learn salient contrasted regions~\cite{Masquelier2007}, which is desirable as these are typically the most informative~\cite{VanRullen2001}. Most of our $C_2$ features turned out to be class-specific, and we could guess what they represent by doing the reconstructions (see Fig.~\ref{features} and Fig. S2). Since each feature results from averaging multiple input images, the specificity of each instance is averaged out, leading to class archetypes. Consequently, good classification results can be obtained using only a few features, or even using `simple' decision rules like feature counts~\cite{Masquelier2007} and majority voting (here), as opposed to a `smart classifier' such as SVM. 

There are some similarities between STDP-based feature learning, and non-negative matrix factorization~\cite{Lee1999}, as first intuited in~\cite{Masquelier2010}, and later demonstrated mathematically in~\cite{Carlson2013}. Within both approaches, objects are represented as (positive) sums of their parts, and the parts are learned by detecting consistently co-active input units.

Our model could be efficiently implemented in hardware, for example using address event representation (AER)~\cite{Zamarreno-Ramos2011,Bichler2012,dorta2016aer,diaz2016efficient}. With AER, the spikes are carried as addresses of sending or receiving neurons on a digital bus. Time `represents itself' as the asynchronous occurrence of the event~\cite{Sivilotti1991}. Thus the use of STDP will lead to a system which effectively becomes more and more reactive, in addition to becoming more and more selective. Furthermore, since biological hardware is known to be incredibly slow, simulations could run several order of magnitude faster than real time~\cite{Serrano-Gotarredona2013}. As mentioned earlier, the primate visual system extracts the rough content of an image in about 100ms. We thus speculate that  some dedicated hardware will  be able to do the same in the order of a millisecond or less.

Recent computational~\cite{Ullman2002}, psychophysical~\cite{harel2011basic}, and fMRI~\cite{lerner2008class} experiments demonstrate that the informative intermediate complexity features are optimal for object categorization tasks. But the possible neural mechanisms to extract such features remain largely unknown. The HMAX model ignores these learning mechanisms and imprints its features with random crops from the training images~\cite{Serre2007.PAMI,Serre2007}, or even uses random filters~\cite{leibo2010primal,Yamins2014a}. Most individual features are thus not very informative, yet in some cases, a `smart' classifier such as  SVM can  efficiently separate the high-dimensional vectors of population responses.

Many other models use supervised learning rules~\cite{LeCun1998,Krizhevsky2012}, sometimes reaching impressive performance on natural image classification tasks~\cite{Krizhevsky2012}. The main drawback of these supervised methods, however, is that learning is slow and requires numerous labeled samples (e.g., about 1 million in~\cite{Krizhevsky2012}), because of the credit assignment problem~\cite{Rolls2002,Ranzato2007}. This contrasts with humans who can generalize efficiently from just a few training examples~\cite{cox2014neural}. We avoid the credit assignment problem by keeping the $C_2$ features fixed when training the final classifier (that being said, fine-tuning them for a given classification problem would probably increase the performance of our model~\cite{Le2013,6763041}; we will test this in future studies). Even if the efficiency of such hybrid unsupervised-supervised learning schemes has been known for a long time, few alternative unsupervised learning algorithms have been shown to be able to extract complex and high-level visual features (see~\cite{Lee2009,Le2013}). Finding better representational learning algorithms is thus an important direction for future research and seeking for inspiration in the biological visual systems is likely to be fruitful~\cite{cox2014neural}. We suggest here that the physiological mechanism known as STDP is an appealing start point.

Considering the time relation among the incoming inputs is an important aspect of spiking neural networks. This property is critical to promote the existing models from static vision to continuous vision~\cite{Masquelier2012}. A prominent example is the trace learning rule~\cite{Foldiak1991}, suggesting that the invariant object representation in ventral visual system is instructed by the implicit temporal contiguity of vision. Also, in various motion processing and action recognition problems~\cite{escobar2009action}, the important information lies in the appearance timing of input features. Our model has this potential to be extended for continuous and dynamic vision -- something that we will further explore.

\section{Conclusions}\label{Conclusion}
To date, various bio-inspired network architectures for object recognition have been introduced, but the learning mechanism of biological visual systems has been neglected. In this paper, we demonstrate that the association of both bio-inspired network architecture and learning rule results in a robust object recognition system. The STDP-based feature learning, used in our model, extracts frequent diagnostic and class specific features that are robust to deformations in stimulus appearance. It has previously been shown that the trivial models can not tolerate the identity preserving transformations such as changes in view, scale, and position. To study the behavior of our model confronted with these difficulties, we have tested our model over two challenging invariant object recognition databases  which includes instances of 10 different object classes photographed in different views, scales, and tilts. The categorization performances indicate that our model is able to robustly recognize objects in such a severe situation. In addition, several analytical techniques have been employed to prove that the main contribution to this success is provided by the unsupervised STDP feature learning, not by the classifier. Using representational dissimilarity matrix, we have shown that the representation of input images in $C_{2}$ layer are more similar for within-category and dissimilar for between-category  objects. In this way, as confirmed by the hierarchical clustering, the objects with the same category are represented in neighboring regions of $C_{2}$ feature space. Hence, even if using a simple classifier, our model is able to reach an acceptable performance, while the random features fail.

\section*{Acknowledgements}
 We would like to thank Mr. Majid Changi Ashtiani at the Math Computing Center of IPM (http://math.ipm.ac.ir/mcc) for letting us to perform some parts of the calculations on their computing cluster. We also thank Dr.~Reza Ebrahimpour for his helpful discussions and suggestions.




%
%
%
\begin{footnotesize}


\end{footnotesize}

\renewcommand\thefigure{S\arabic{figure}}
\setcounter{figure}{0}

\onecolumn 
\section*{Supplementary Information}
\begin{figure*}[htb]
\centering
\includegraphics[scale=0.6]{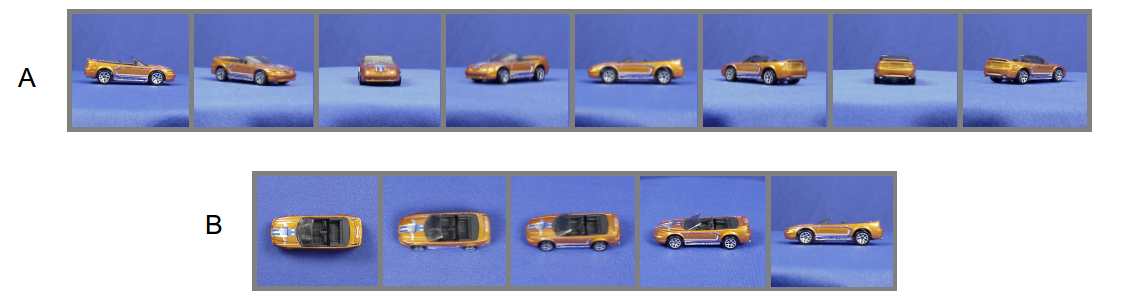}
\caption{Some images of car class of ETH-80 dataset in different A) views, and B) tilts.}
\label{ETH_samples}
\end{figure*}

\begin{figure*}
\centering
\subfloat[View 0\degree]{\includegraphics[width=1.7in]{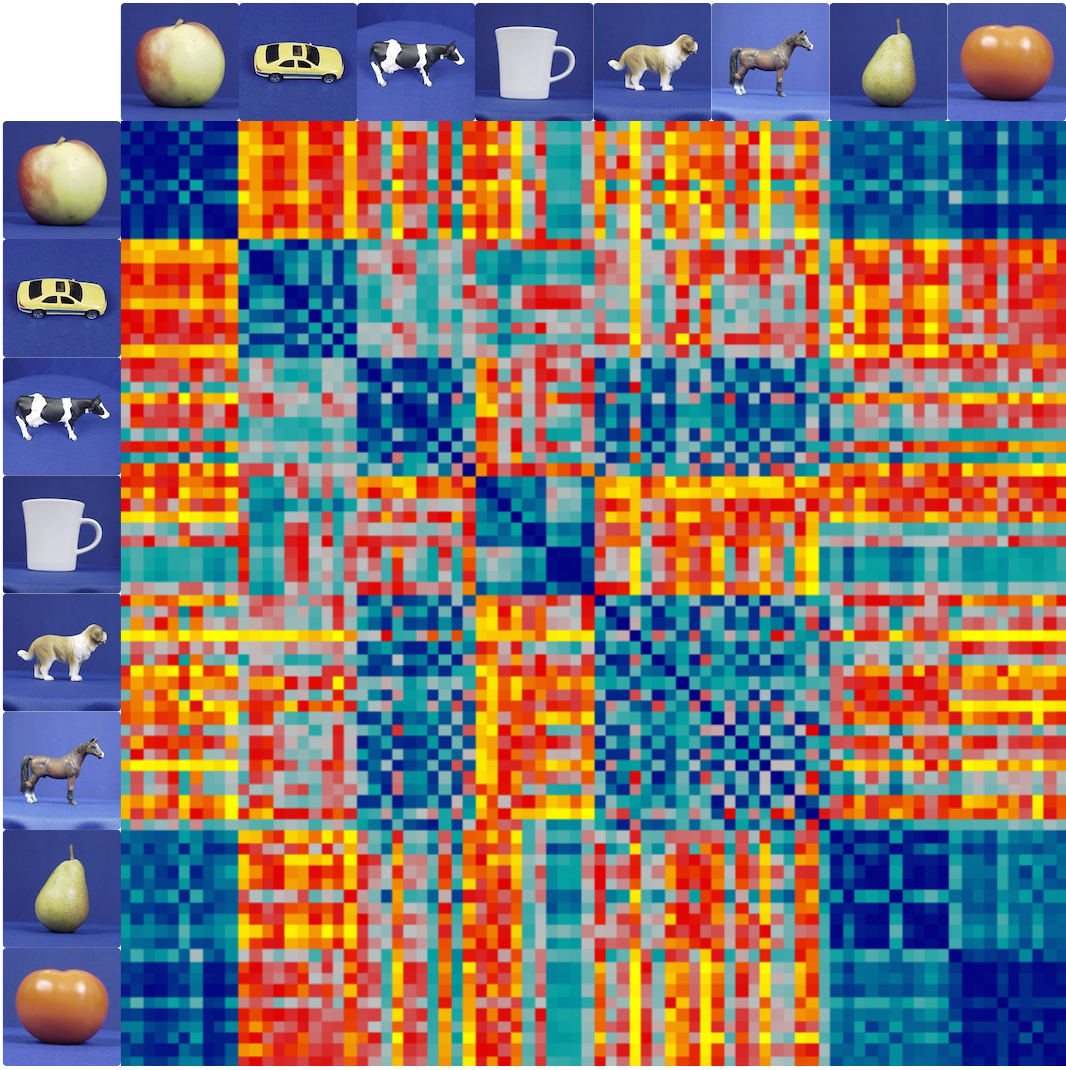}%
\label{fig_supliment_first_case2}}
\hfil
\subfloat[View 45\degree]{\includegraphics[width=1.7in]{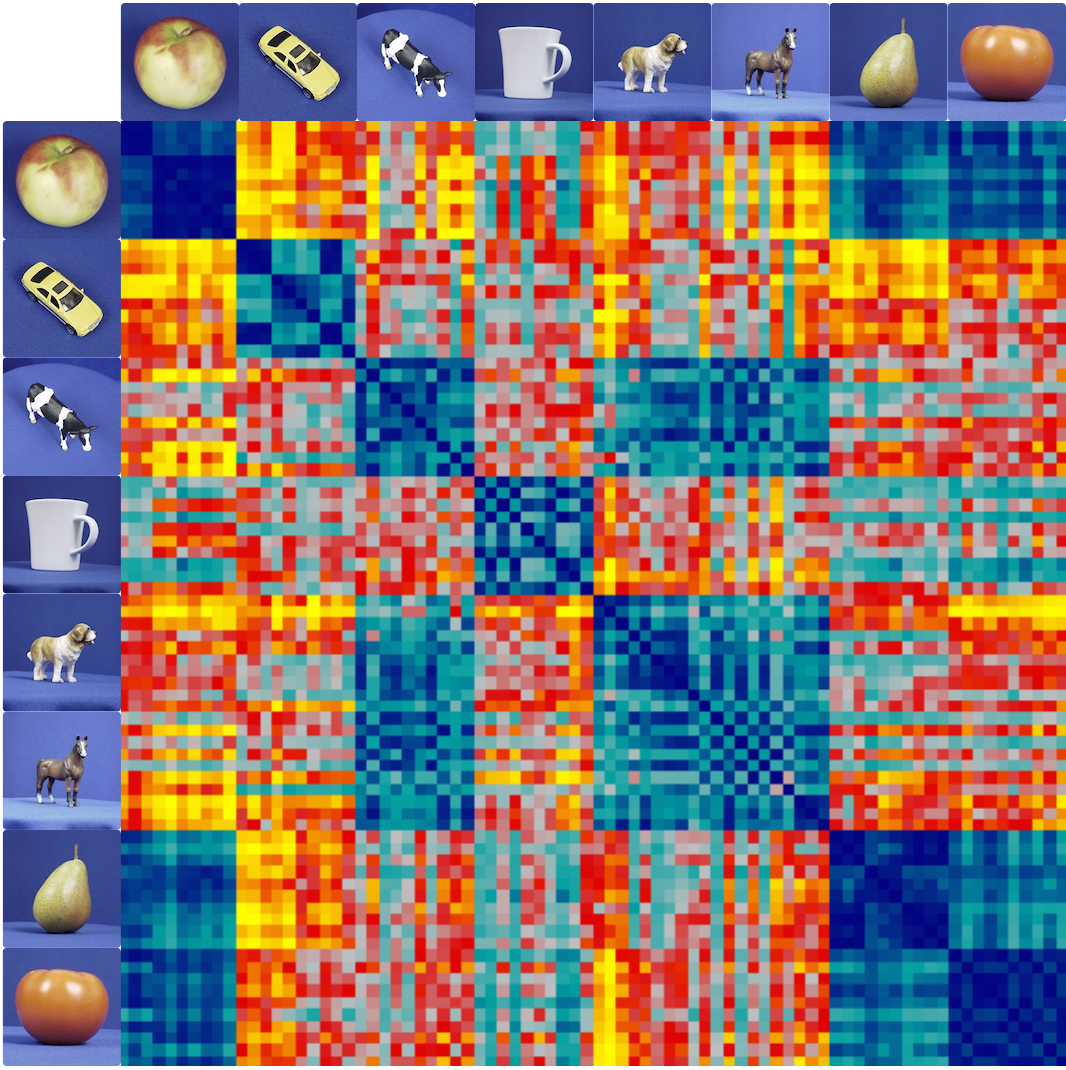}%
\label{fig_supliment_second_case2}}
\hfil
\subfloat[view 90\degree]{\includegraphics[width=1.7in]{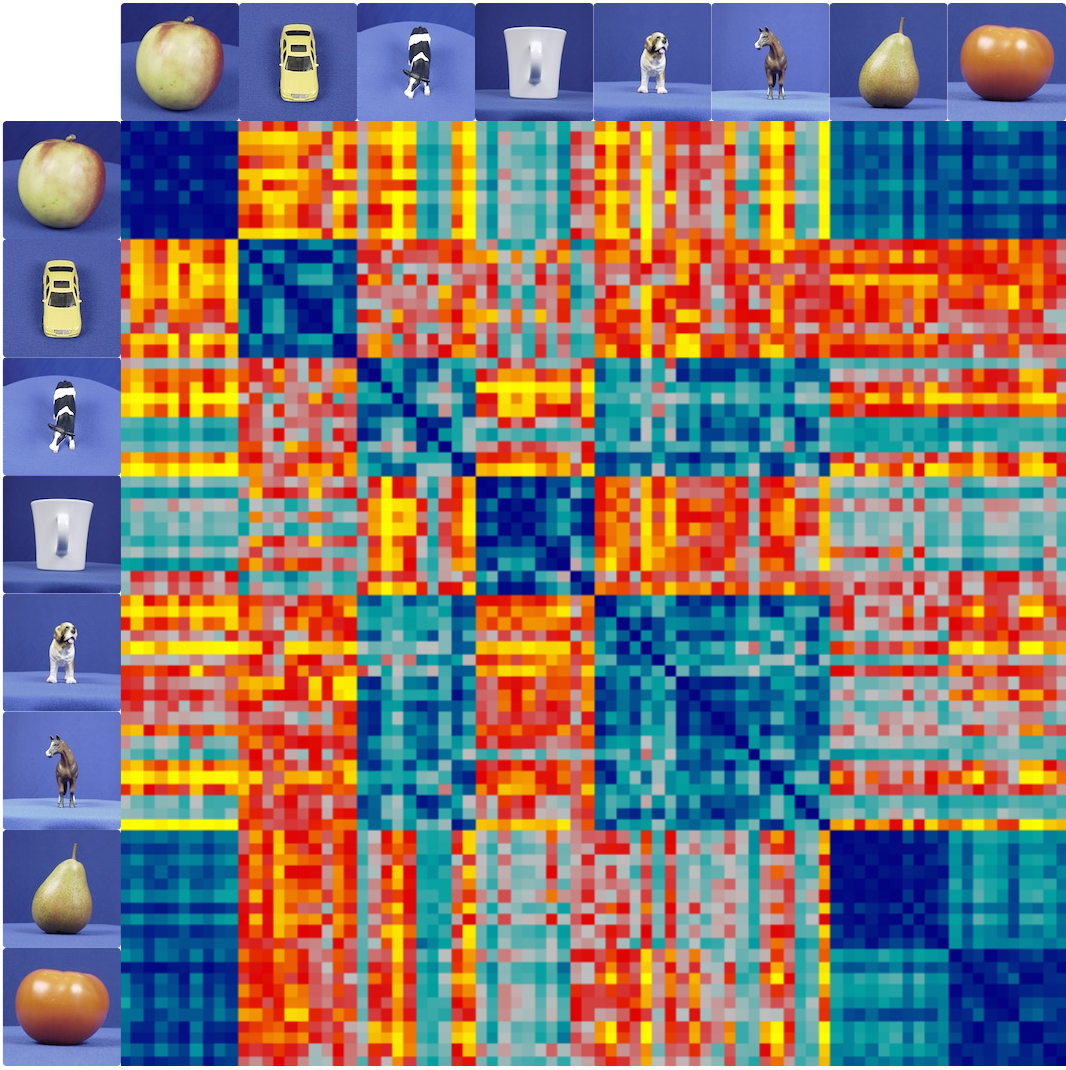}%
\label{fig_supliment_3rd_case3}}
\hfil
\subfloat[view 135\degree]{\includegraphics[width=1.7in]{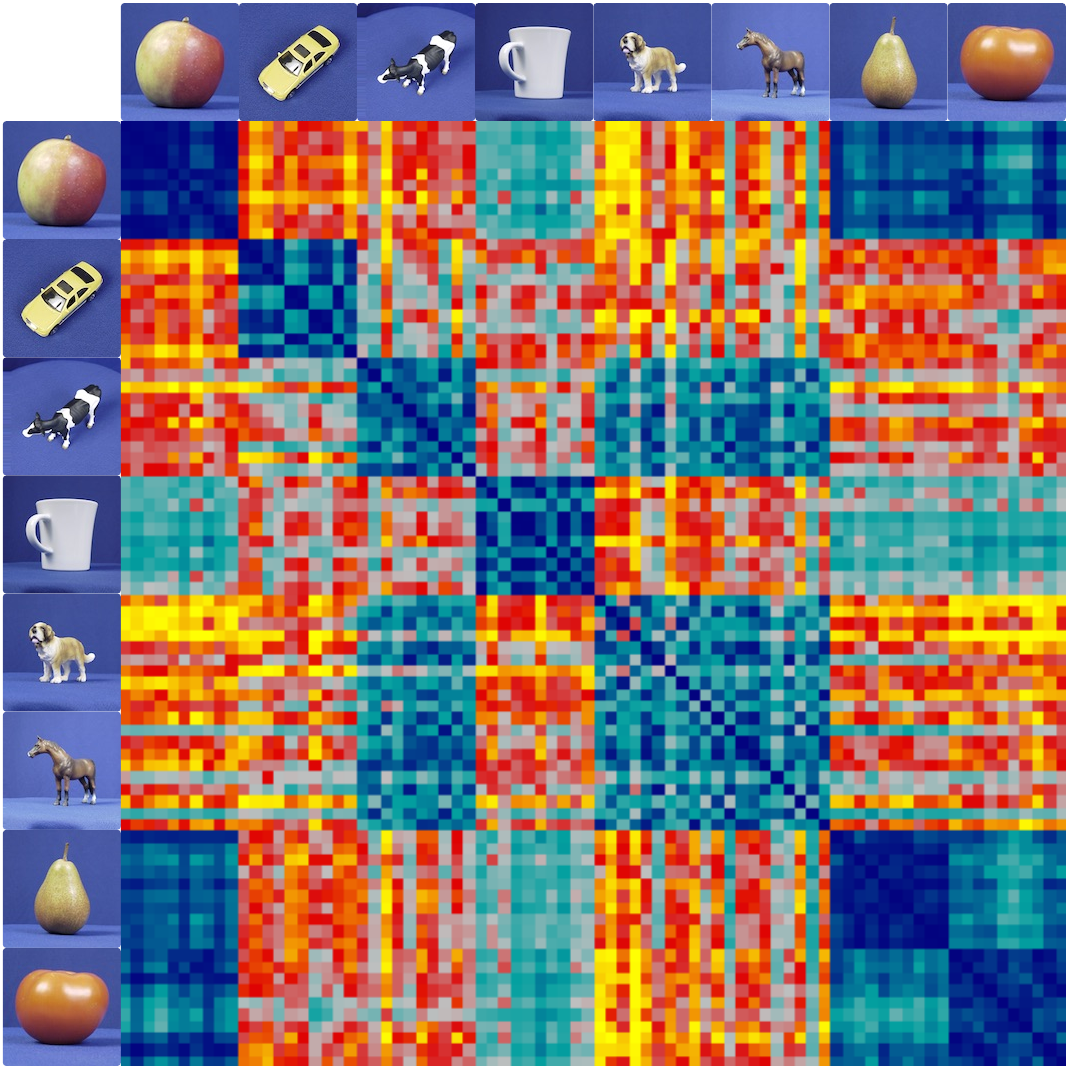}%
\label{fig_supliment_4th_case2}}
\hfil
\subfloat[view 180\degree]{\includegraphics[width=1.7in]{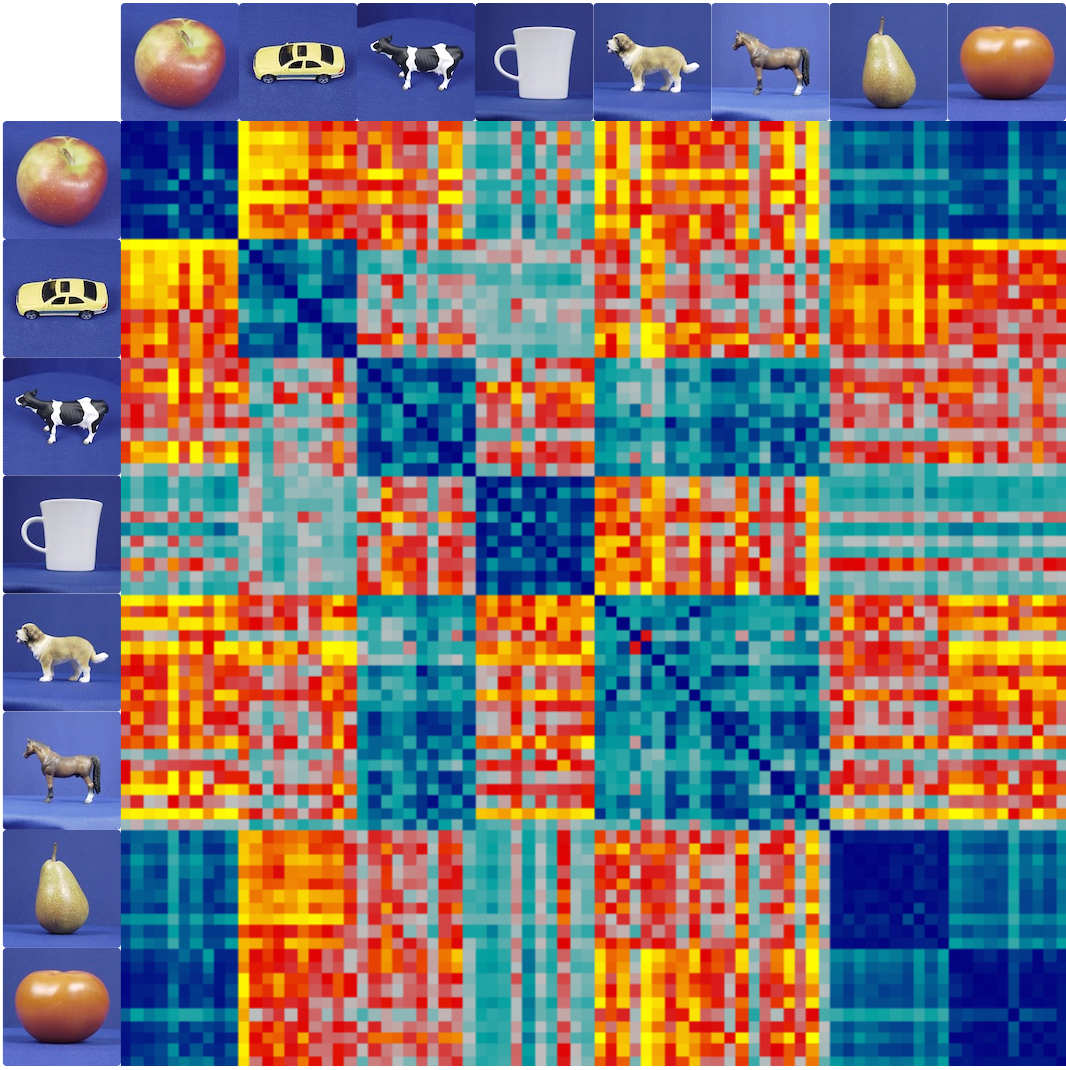}%
\label{fig_supliment_5th_case2}}
\hfil
\subfloat[View 225\degree]{\includegraphics[width=1.7in]{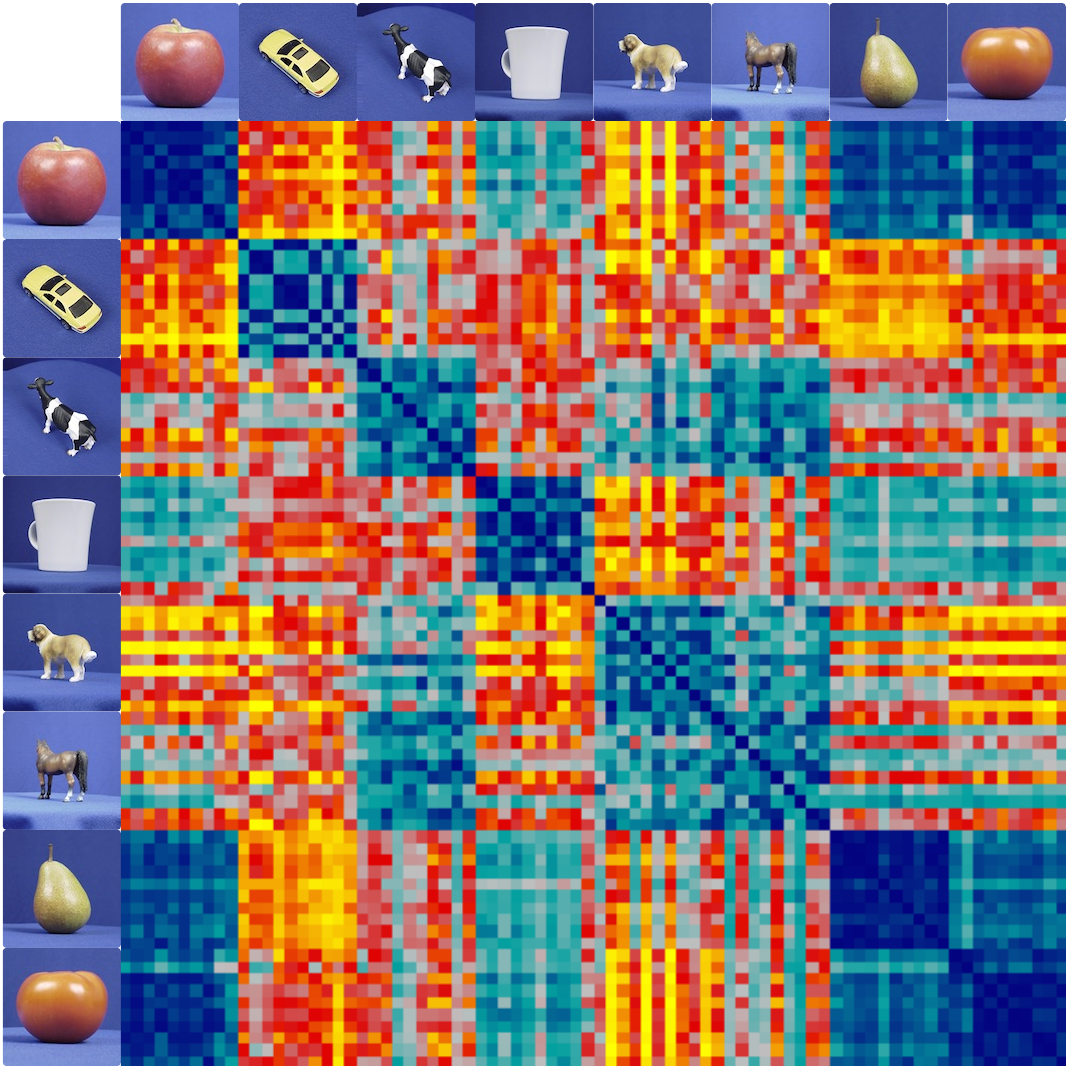}%
\label{fig_supliment_6th_case2}}
\hfil
\subfloat[View 270\degree]{\includegraphics[width=1.7in]{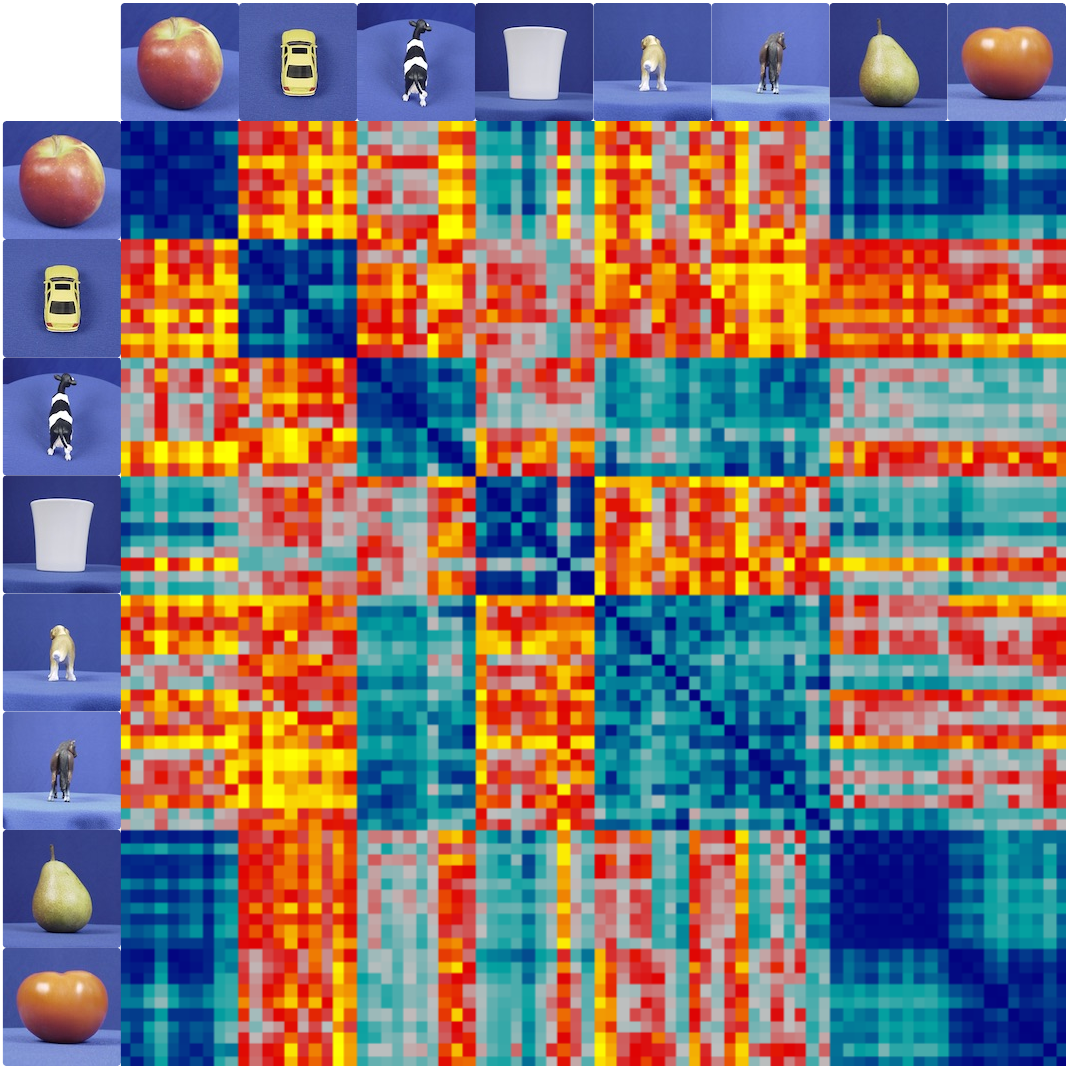}%
\label{fig_supliment_7th_case2}}
\hfil
\subfloat[View 315\degree]{\includegraphics[width=1.7in]{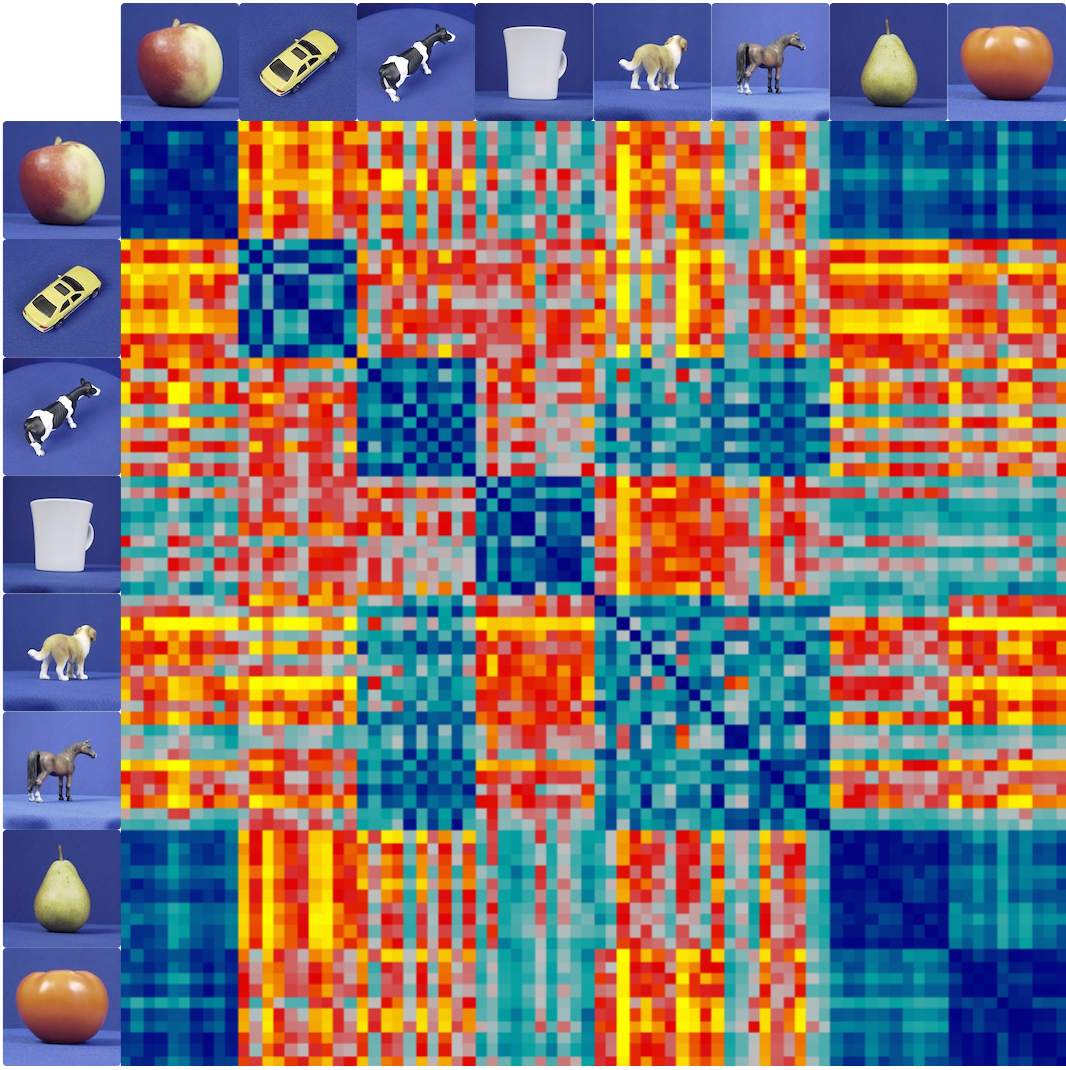}%
\label{fig_supliment_8th_case2}}
\hfil
\vspace*{0.5cm}
\includegraphics[width=2.2in]{Legend}%

\caption{RDMs for C2 features our model on ETH-80 corresponding to different viewpoints.}
\label{RDMs_supliment_Tims}
\end{figure*}

\begin{figure*}
\centering
\subfloat[View 0\degree]{\includegraphics[width=1.7in]{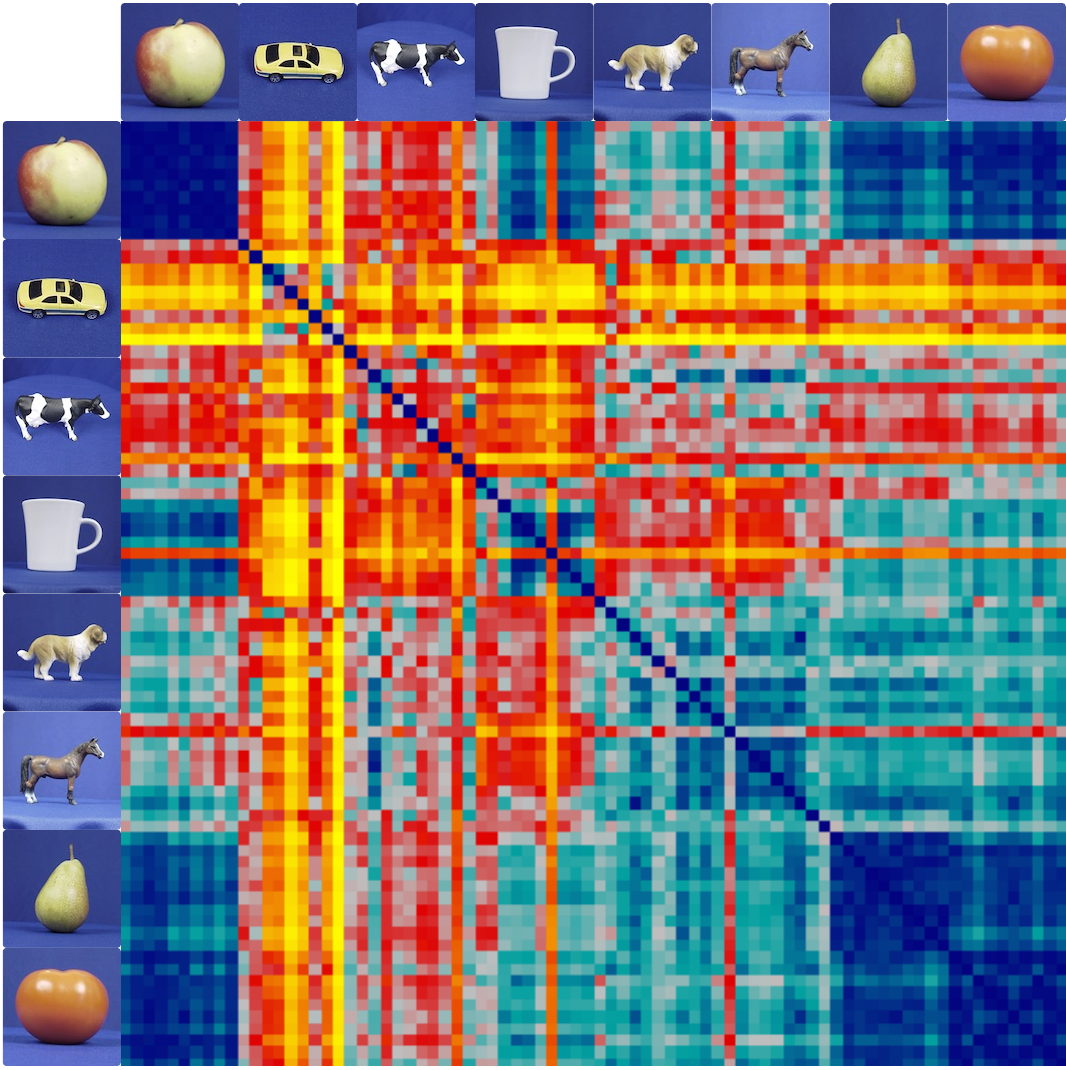}%
\label{fig_supliment_first_case1}}
\hfil
\subfloat[View 45\degree]{\includegraphics[width=1.7in]{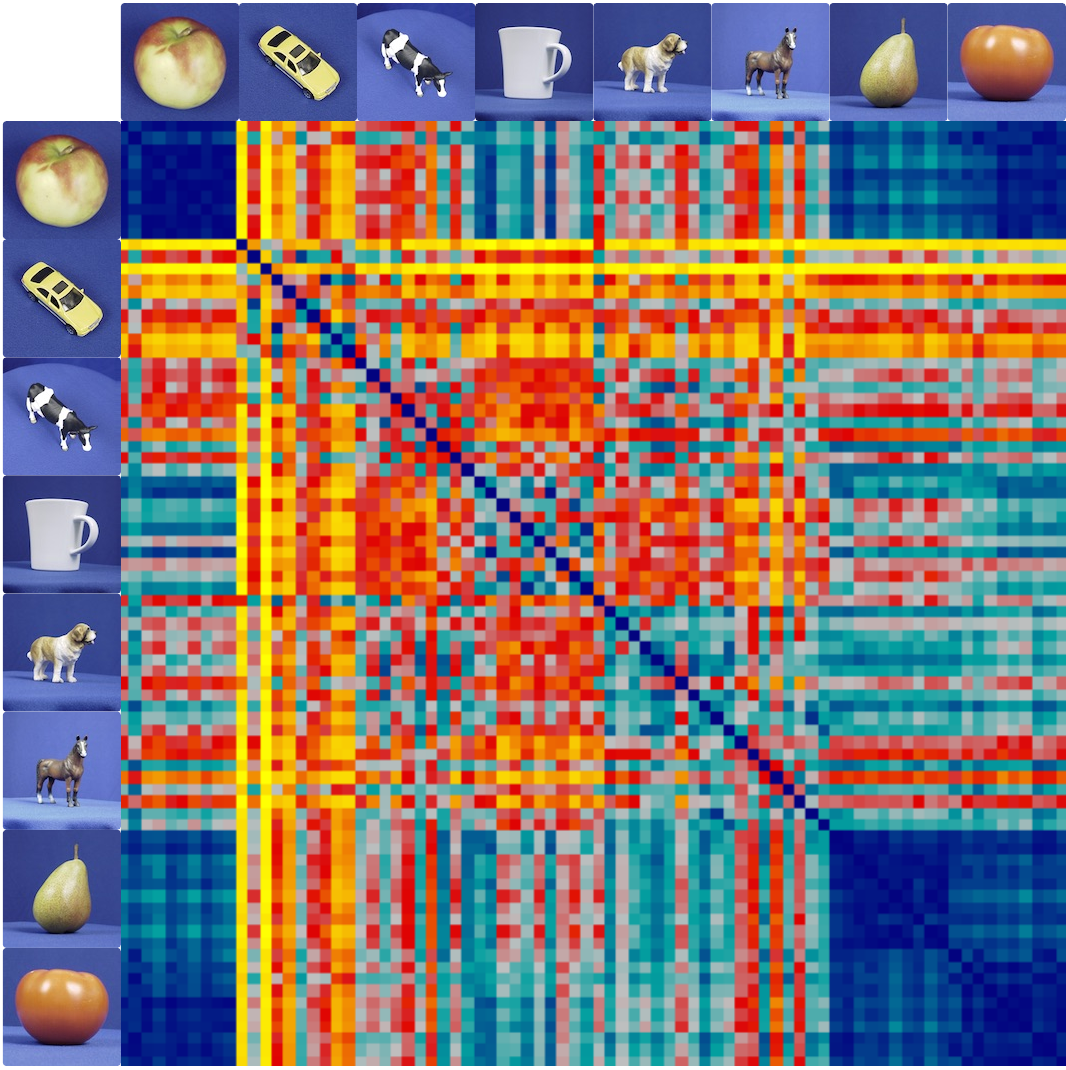}%
\label{fig_supliment_second_case1}}
\hfil
\subfloat[view 90\degree]{\includegraphics[width=1.7in]{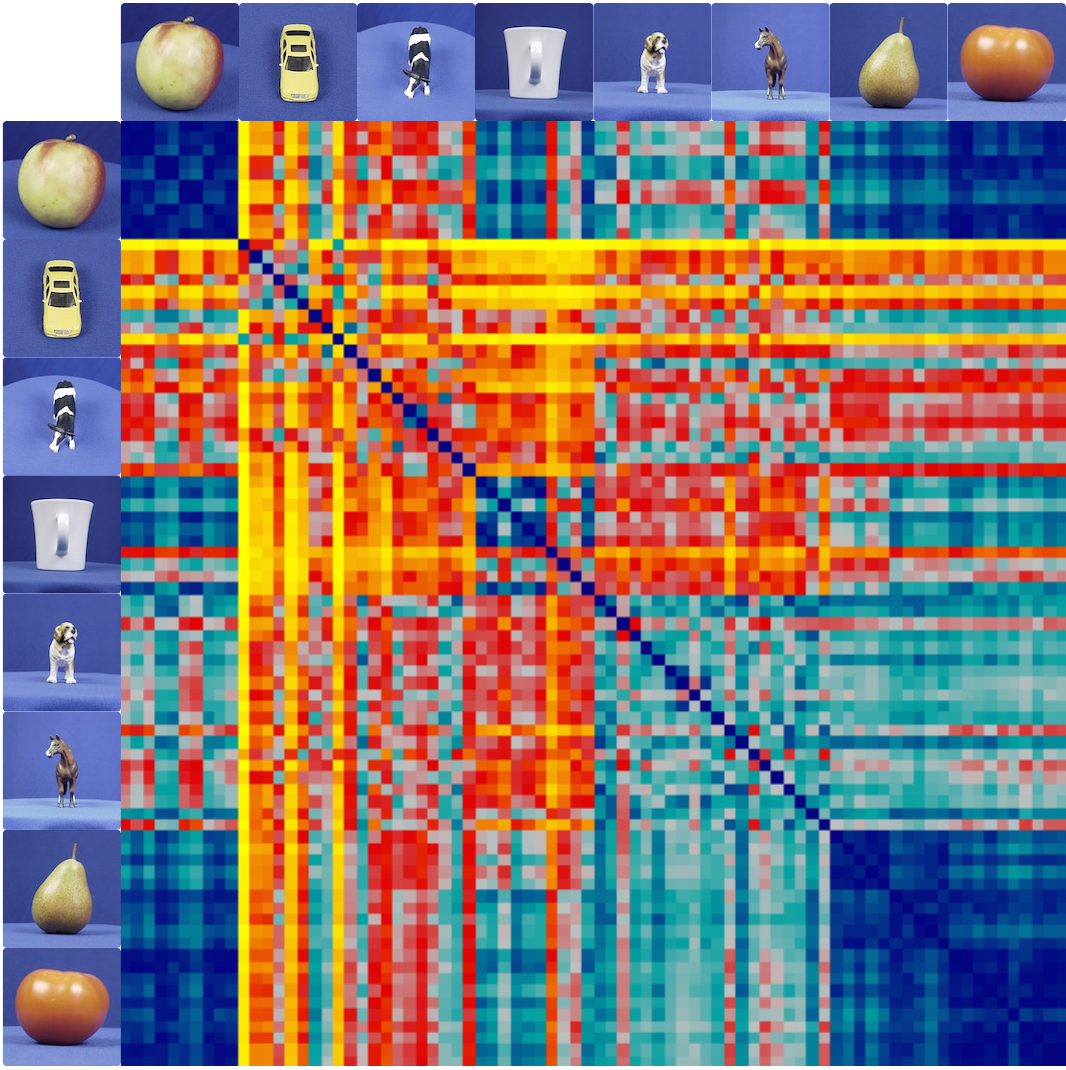}%
\label{fig_supliment_3rd_case1}}
\hfil
\subfloat[view 135\degree]{\includegraphics[width=1.7in]{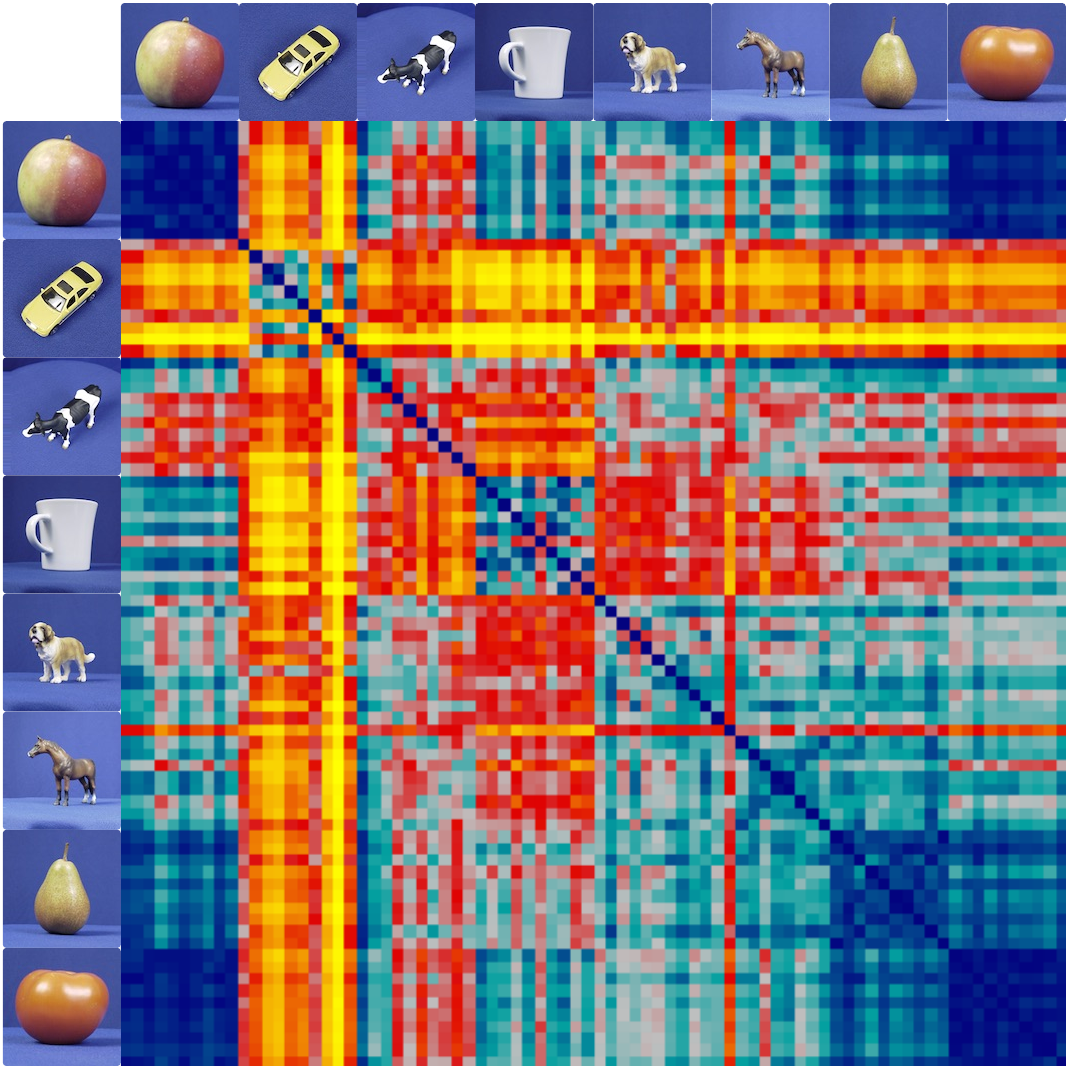}%
\label{fig_supliment_4th_case1}}
\hfil
\subfloat[view 180\degree]{\includegraphics[width=1.7in]{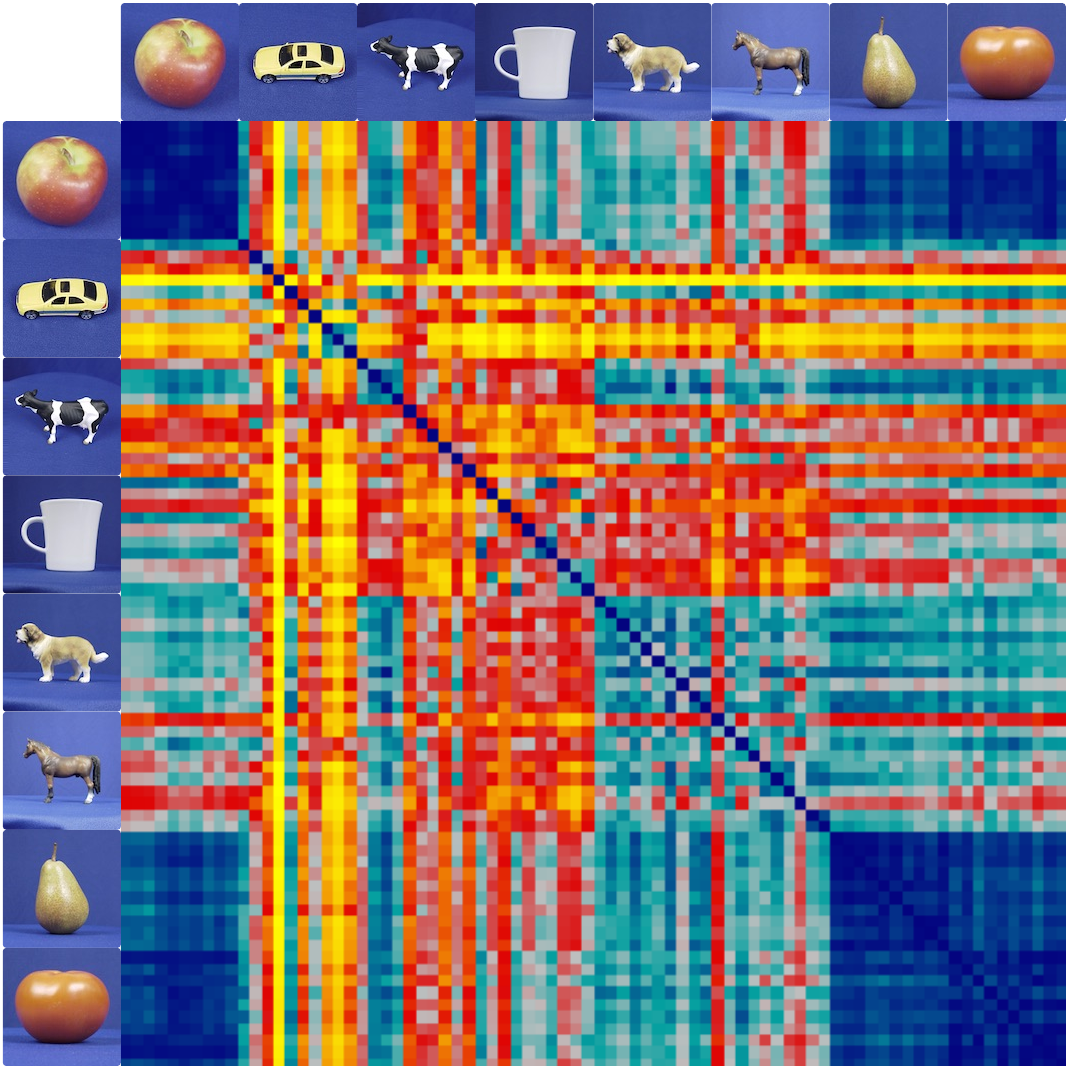}%
\label{fig_supliment_5th_case1}}
\hfil
\subfloat[View 225\degree]{\includegraphics[width=1.7in]{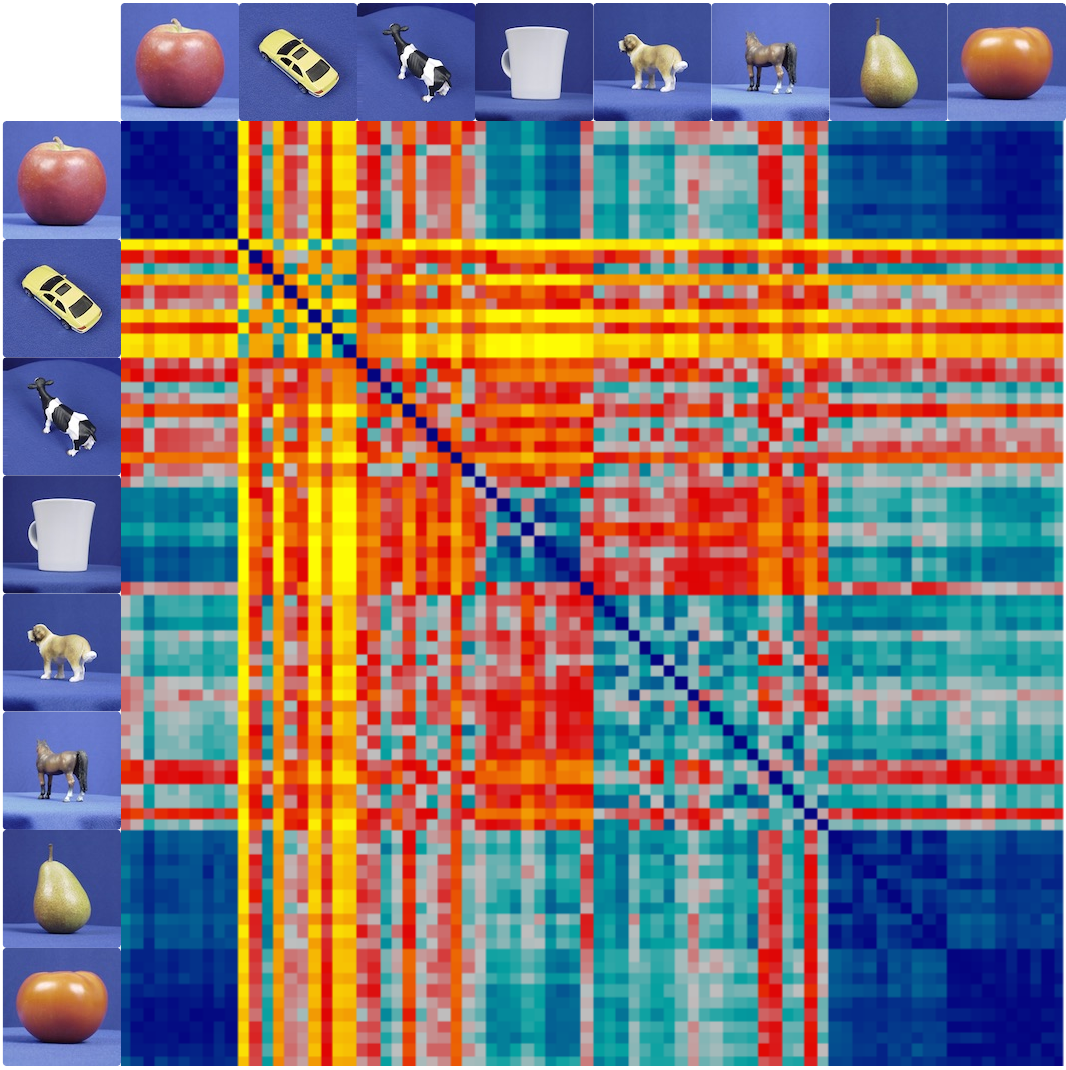}%
\label{fig_supliment_6th_case1}}
\hfil
\subfloat[View 270\degree]{\includegraphics[width=1.7in]{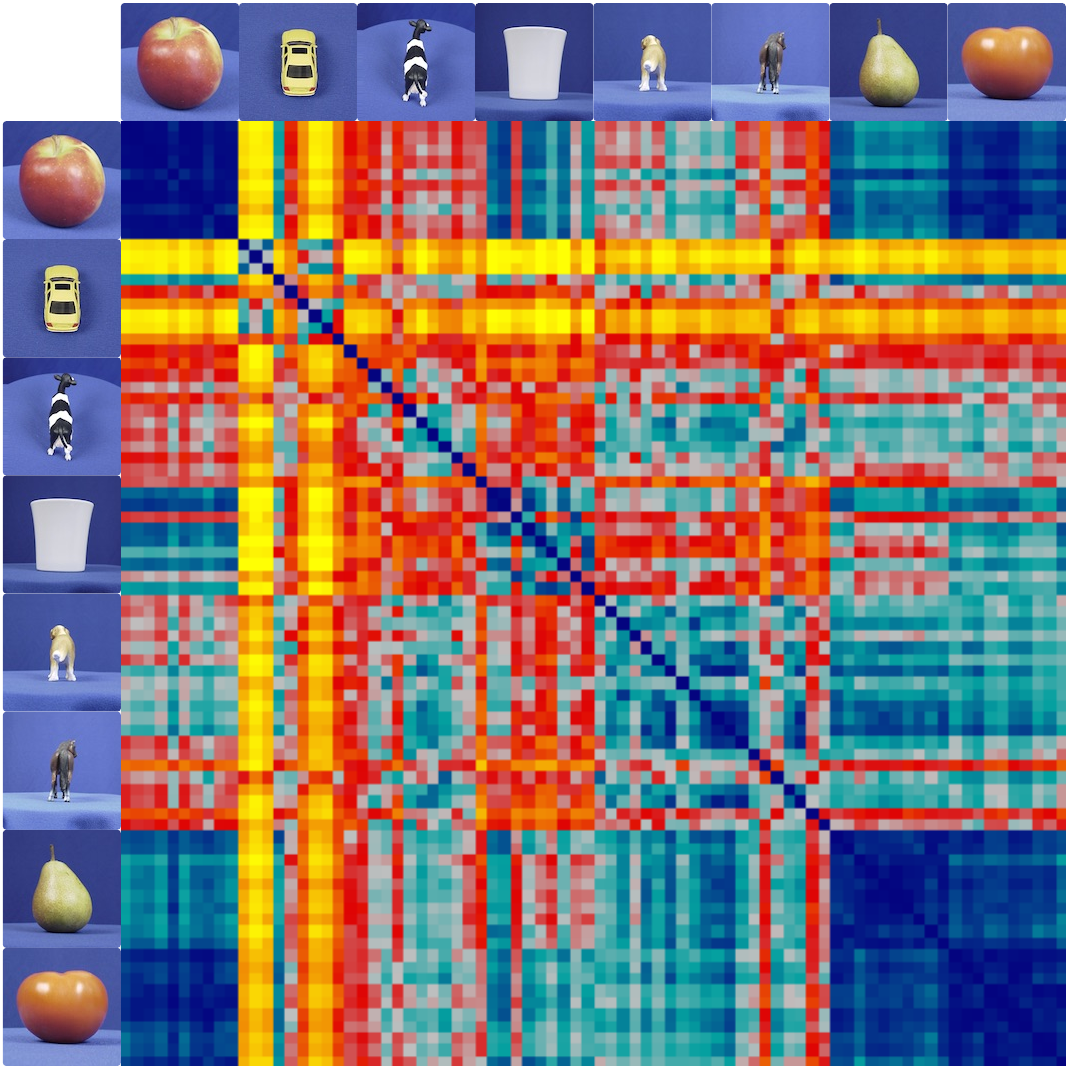}%
\label{fig_supliment_7th_case1}}
\hfil
\subfloat[View 315\degree]{\includegraphics[width=1.7in]{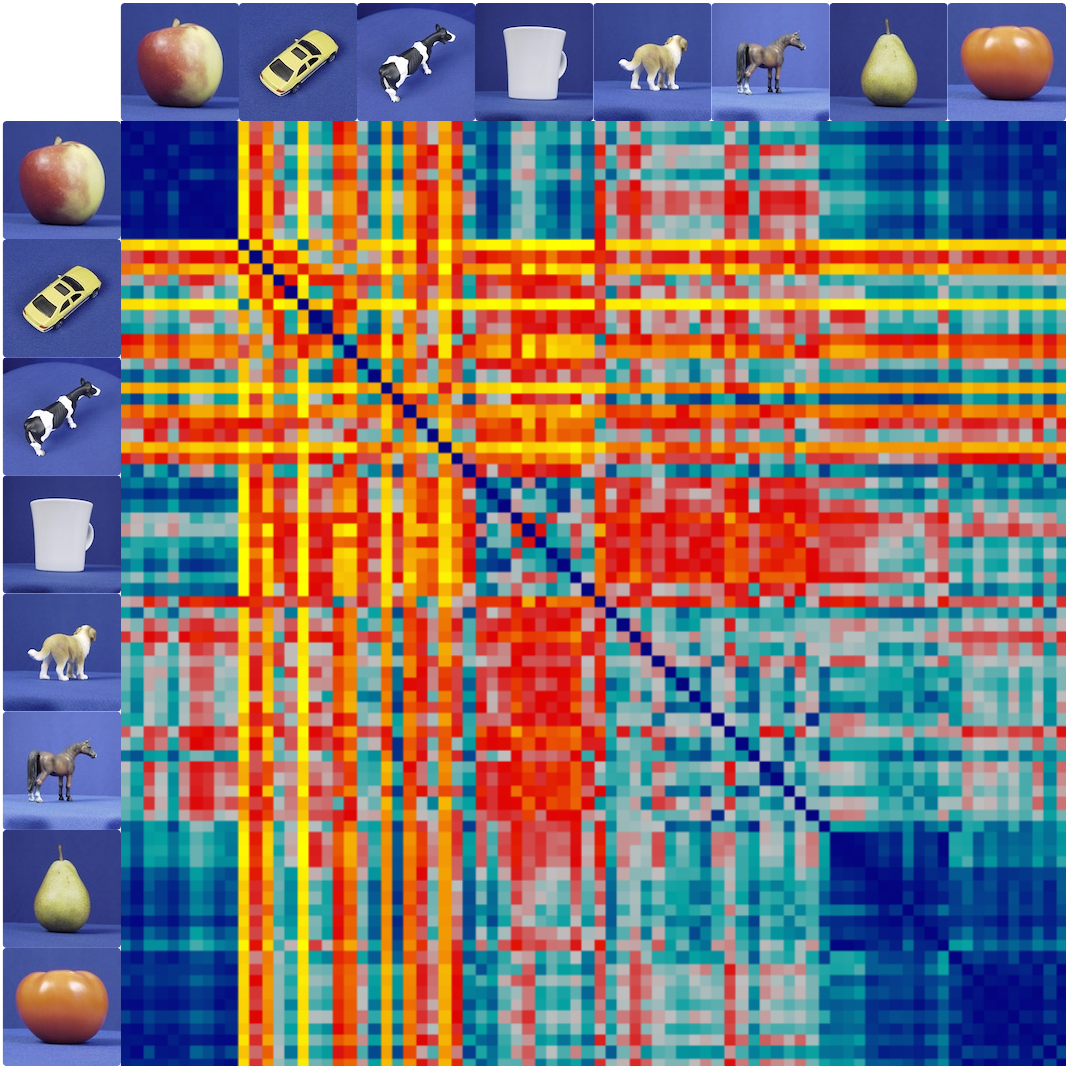}%
\label{fig_supliment_8th_case1}}
\hfil
\vspace*{0.5cm}
\includegraphics[width=2.2in]{Legend}%

\caption{RDMs of HMAX C2 features on ETH-80 corresponding to different viewpoints.}
\label{RDMs_supliment_HMAX}
\end{figure*}

\begin{figure*}
\centering
\includegraphics[scale=0.75]{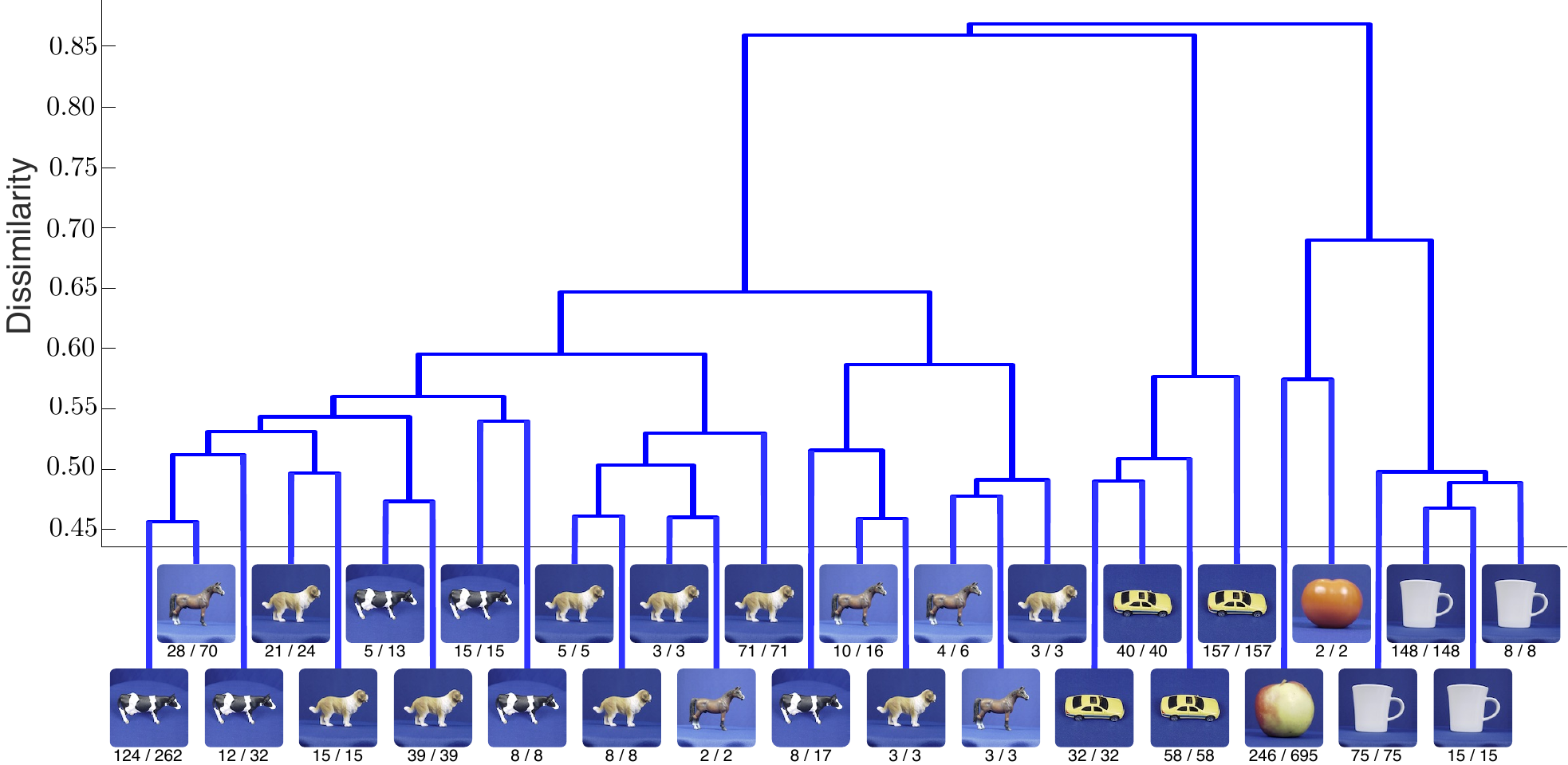}
\caption{The hierarchy of clusters and their labels for our model on ETH-80 dataset. The label of each cluster indicates the class with the highest frequency in that cluster. The cardinality of each cluster, $C$, and the cardinality of the class with the highest frequency, $H$, are placed below the cluster label as $H/C$.}
\label{Cluster_TIM_ETH80}
\end{figure*}

\begin{figure*}
\centering
\includegraphics[scale=0.75]{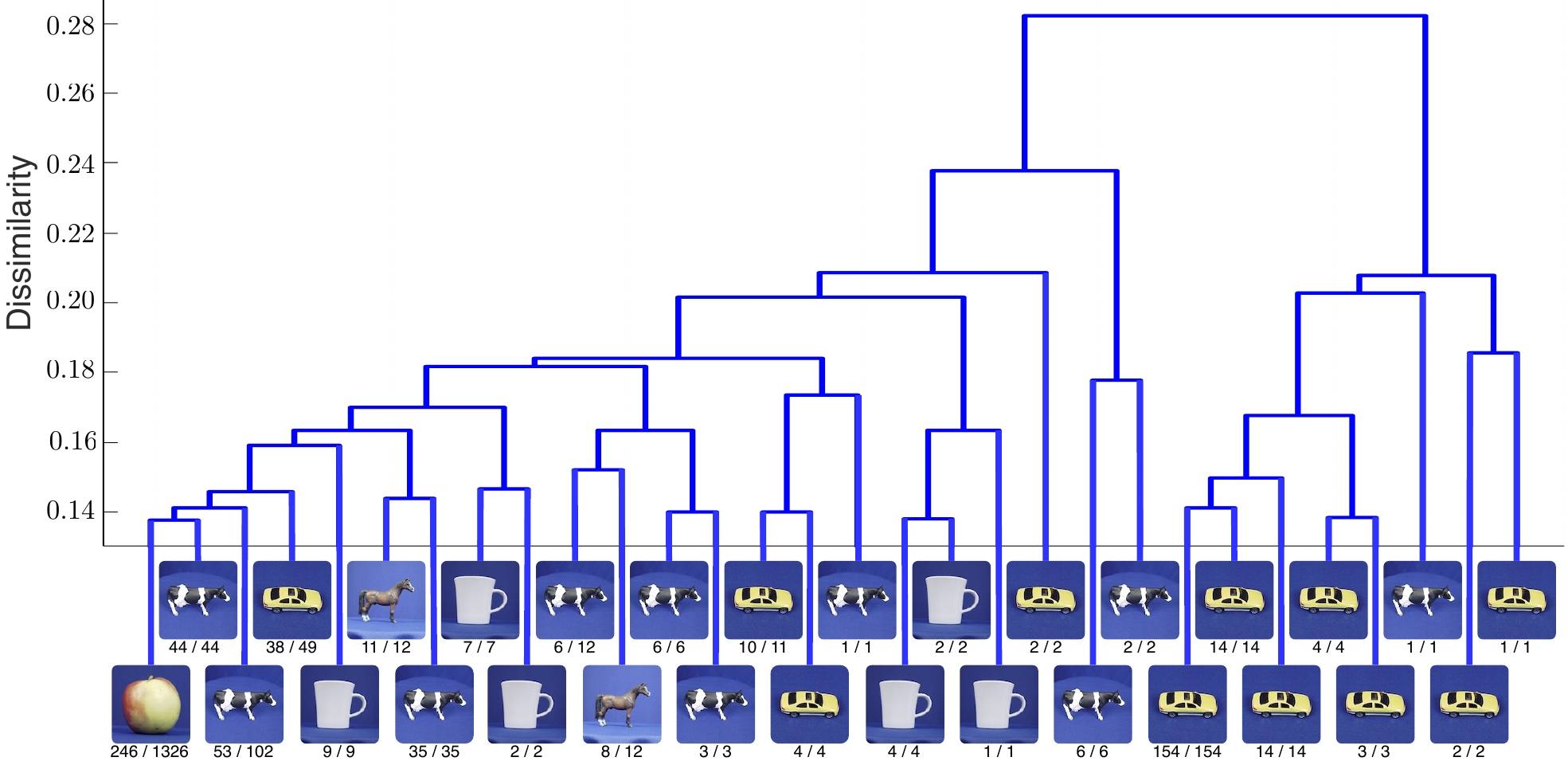}
\caption{The hierarchy of clusters and their labels for the HMAX model on ETH-80 dataset. The label of each cluster indicates the class with the highest frequency in that cluster. The cardinality of each cluster, $C$, and the cardinality of the class with the highest frequency, $H$, are placed below the cluster label as $H/C$.}
\label{Cluster_HMax_ETH80}
\end{figure*}

Here we provide the results of feature analysis techniques such as RDM and hierarchical clustering on ETH-80 dataset for both HMAX and our model. Some sample images of ETH-80 dataset are shown in Fig.~\ref{ETH_samples}. In Fig.~\ref{RDMs_supliment_Tims} and Fig.~\ref{RDMs_supliment_HMAX} the RDMs of $C2$ features of our model and HMAX in eight view angels are presented, respectively. It can be seen that our model can better represent classes with high shape similarities such as tomato, apple, and pear or cow, horse, and dog with respect to the HMAX model. Also, the hierarchical clustering of whole training data based on their representations on feature spaces of our model and   HMAX  are demonstrated in Fig.~\ref{Cluster_TIM_ETH80} and Fig.\ref{Cluster_HMax_ETH80}, respectively. As for the 3D-Object dataset, HMAX feature extraction leads to a nested representation of different object classes which causes a poor classification accuracy. Here again a huge number of images which belong to different classes are assigned to a large cluster with lower than 0.14 internal dissimilarities. On the other hand, our model has distributed images of different classes in different regions of $C2$ feature space. Note that the largest cluster of our model includes the instances of tomato, apple, and pear classes which their shapes are so similar.

\begin{figure*}
\centering
\subfloat[]{\includegraphics[width=2.2in]{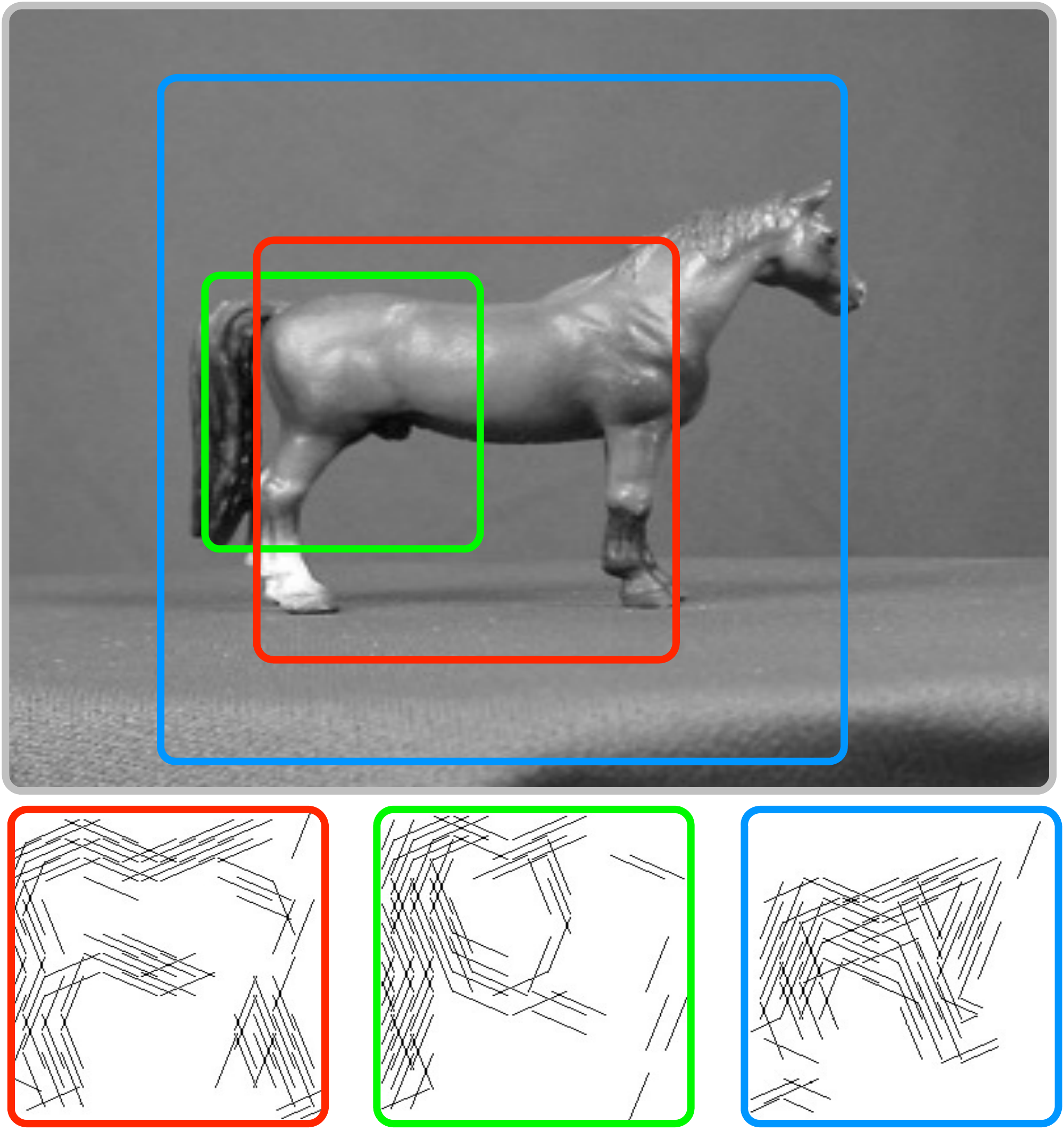}%
\label{fig_first_case}}
\hfil
\subfloat[]{\includegraphics[width=2.2in]{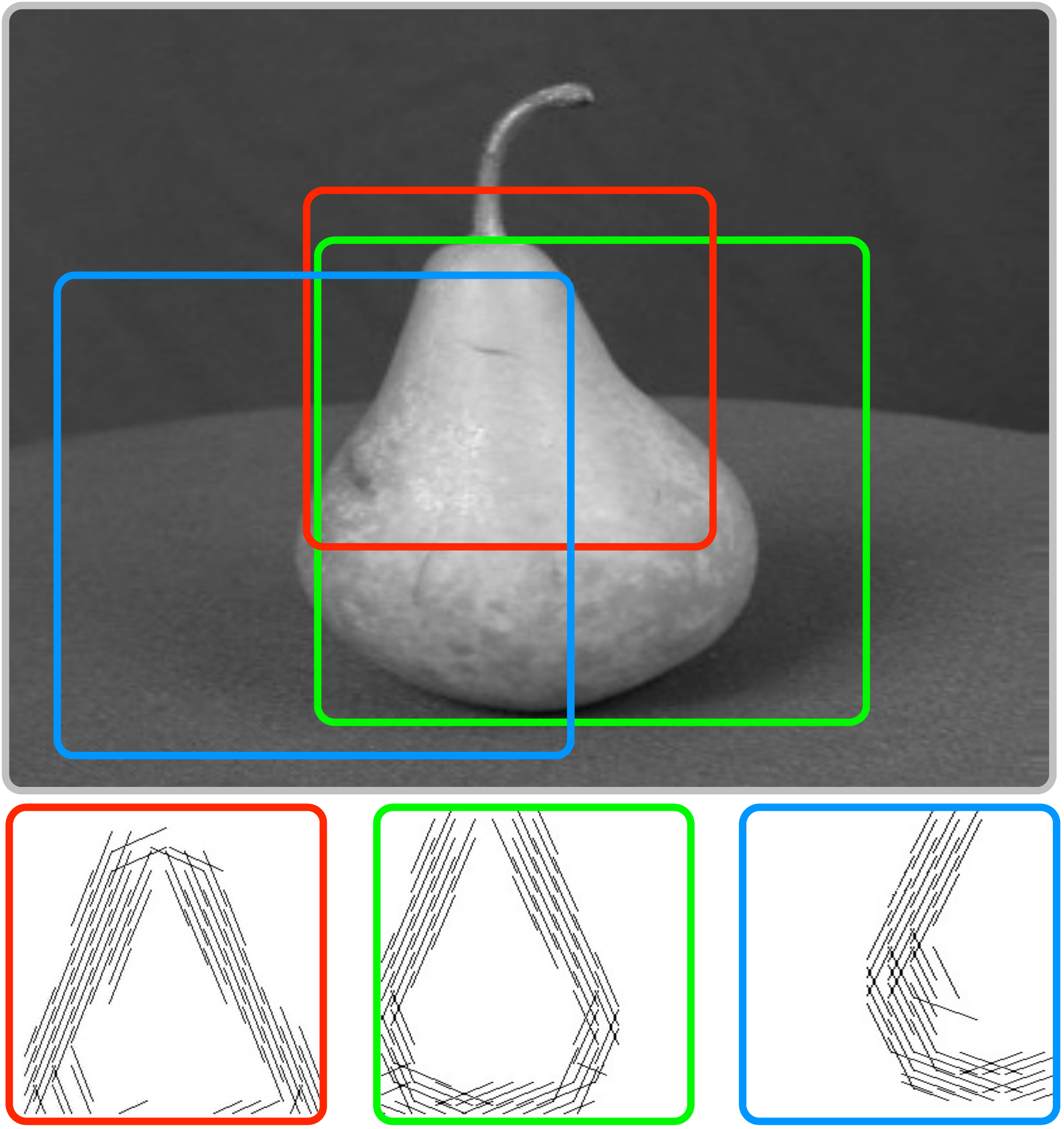}%
\label{fig_second_case}}
\hfil
\subfloat[]{\includegraphics[width=2.2in]{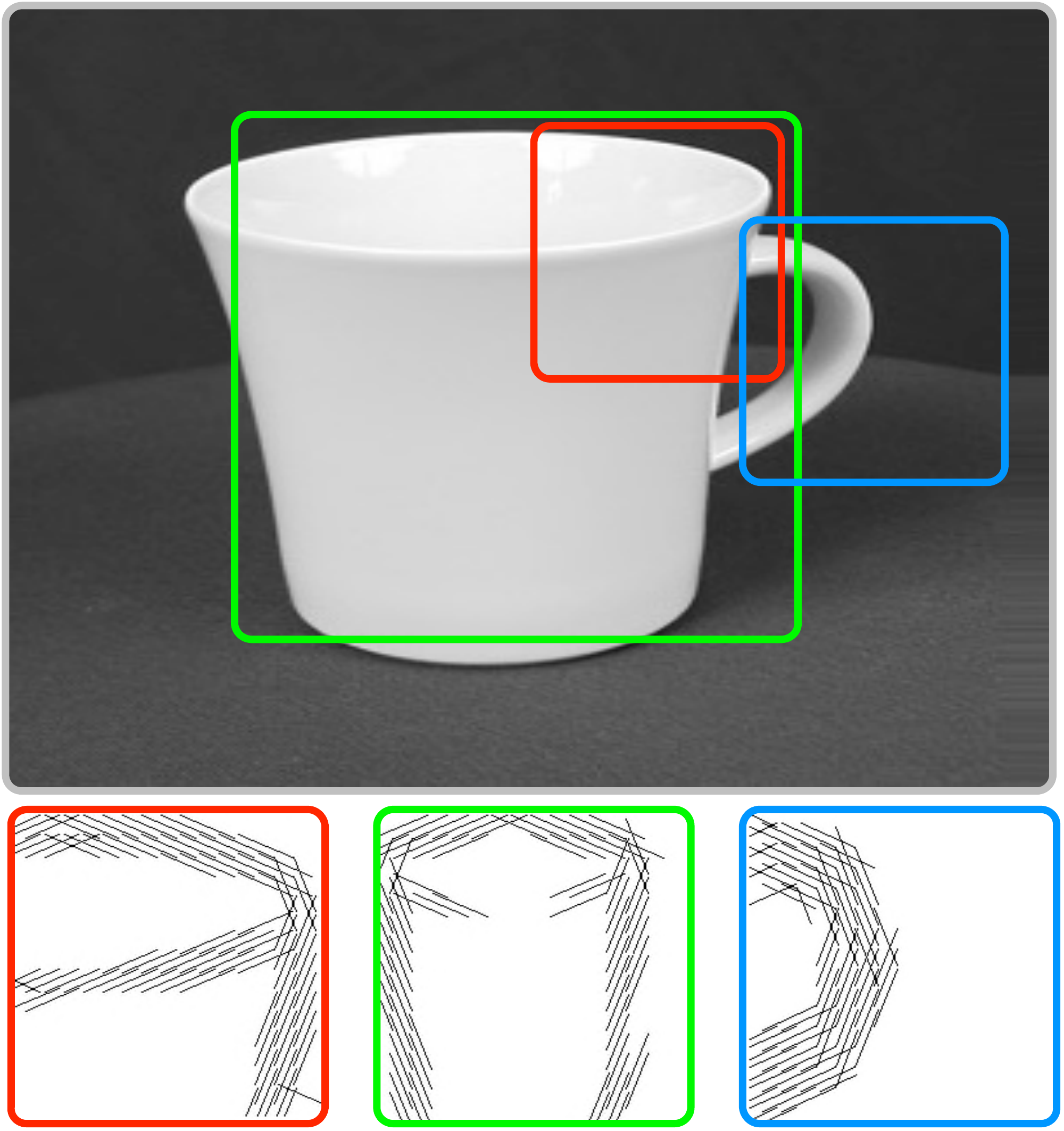}%
\label{fig_3rd_case}}
\caption{Three $S_2$ feature prototypes selective to the a) horse, b) pear, and c) cup classes of ETH-80 dataset along with their reconstructed preferred stimuli.}
\label{features}
\end{figure*}


\end{document}